\documentclass{egpubl}
\usepackage{egsr2025_DLOnlyTrack}
 
\WsPaper           %
\CGFccby

\usepackage[T1]{fontenc}
\usepackage{dfadobe}  
\usepackage[normalem]{ulem}
\usepackage{cite}  %
\BibtexOrBiblatex
\electronicVersion
\PrintedOrElectronic
\ifpdf \usepackage[pdftex]{graphicx} \pdfcompresslevel=9
\else \usepackage[dvips]{graphicx} \fi

\usepackage{egweblnk} 
\usepackage{mathtools} 
\usepackage[capitalize]{cleveref}
\DeclareUnicodeCharacter{2194}{\ensuremath{\leftrightarrow}}

\usepackage[table,xcdraw]{xcolor}
\usepackage{xspace} 
\usepackage{xcolor}
\usepackage{subcaption}

\newcommand{\FLIP}{\protect\reflectbox{F}LIP\xspace}
\newcommand{\bgcolor}[2]{\setlength{\fboxsep}{0pt}\colorbox{#1}{\strut #2}}
\newcommand{\BGcolor}[3][HTML]{\definecolor{mycolor}{HTML}{#2}\bgcolor{mycolor}{#3}}

\title[An evaluation of SVBRDF Prediction from Generative Image Models for Texturing 3D Scenes]%
      {An evaluation of SVBRDF Prediction from Generative Image Models for Appearance Modeling of 3D Scenes}

\author[A. Gauthier et al.]
{\parbox{\textwidth}{\centering 
        A. Gauthier$^{1}$\orcid{0000-0002-3710-0879},
        V. Deschaintre$^{2}$\orcid{0000-0002-6219-3747},
        A. Lanvin$^{1}$\orcid{0009-0003-5343-2528},
        F. Durand$^{3}$\orcid{0000-0001-9919-069X},
        A. Bousseau$^{1}$\orcid{0000-0002-8003-9575},
        and G. Drettakis$^{1}$\orcid{0000-0002-9254-4819} 
        }
        \\
{\parbox{\textwidth}{\centering $^1$Inria \& Université Côte d’Azur, France
         $^2$Adobe Research, UK
         $^3$MIT, USA\\
       }
}
\\ \\
\small{Project \& Code:} \href{https://repo-sam.inria.fr/nerphys/svbrdf-evaluation/}{repo-sam.inria.fr/nerphys/svbrdf-evaluation}
}

\begin{document}

\teaser{
 \includegraphics[width=1.0\linewidth]{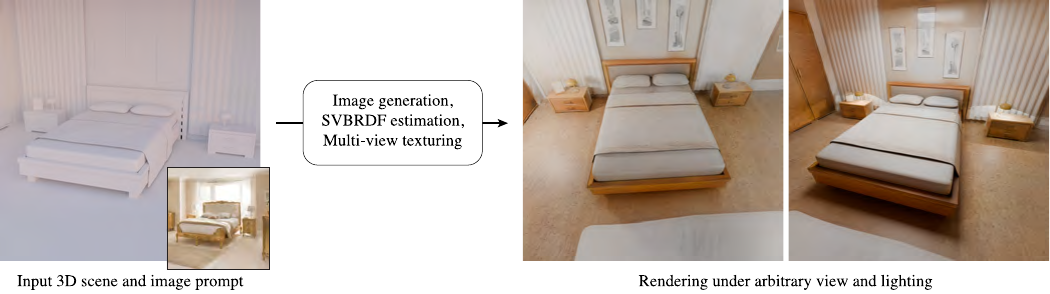}
 \centering
  \caption{
    We evaluate design choices for implementing a fast pipeline to model appearance of 3D scene assets. Given the geometry of the scene and an example image (left), we use generative image diffusion models and SVBRDF predictors to obtain multiview physically-based material maps that are merged into a texture atlas for the scene, enabling rendering under arbitrary view and lighting (right, see accompanying video for animated results). We evaluate how different single-image SVBRDF predictors -- originally developed to process photographs -- perform in this generative context, both in terms of per-view accuracy and in terms of multiview coherence. 
    }
\label{fig:teaser}
}

\maketitle
\begin{abstract}
    Digital content creation is experiencing a profound change with the advent of deep generative models. For texturing, conditional image generators now allow the synthesis of realistic RGB images of a 3D scene that align with the geometry of that scene. For appearance modeling, SVBRDF prediction networks recover material parameters from RGB images. Combining these technologies allows us to quickly generate SVBRDF maps for multiple views of a 3D scene, which can be merged to form a SVBRDF texture atlas of that scene. In this paper, 
    we analyze the challenges and opportunities for SVBRDF prediction in the context of such a fast appearance modeling pipeline.
    On the one hand, single-view SVBRDF predictions might suffer from multiview incoherence and yield inconsistent texture atlases. On the other hand, generated RGB images, and the different modalities on which they are conditioned, can provide additional information for SVBRDF estimation compared to photographs. We compare neural architectures and conditions to identify designs that achieve high accuracy and coherence. We find that, surprisingly, a standard UNet is competitive with more complex designs.

\ccsdesc[500]{Computing methodologies~Texturing}
\ccsdesc[500]{Computing methodologies~Reflectance modeling}

\printccsdesc   
\end{abstract}

\section{Introduction}

Modeling appearance for 3D scenes is one of the hardest steps in 3D content creation, requiring painstaking manual selection and tweaking of material maps that encode the spatially-varying parameters of BRDFs (albedo, roughness, metallic).
While generative diffusion models have demonstrated potential for the related tasks of image creation \cite{song2021denoising,ho2020denoising} and texturing \cite{Richardson2023TEXTure,chen2023text2tex}, the very large datasets of images they require for training specialize them to the RGB domain. In parallel, the field of single-image material estimation has matured to offer solutions for estimating SVBRDFs at scene scale \cite{li2020inverse,kocsis2024iid,RGB2X2024Zeng}.
We observe that image generators can be combined with material estimators to form a fast pipeline for appearance modeling of 3D scenes (Figure~\ref{fig:teaser}).
In this paper, we study the design choices that underpin such a pipeline and compare different single-image SVBRDF predictors when applied to multiple generated images of a scene, seeking for both accurate per-view estimation and coherence over multiple views, since these two properties are critical for merging multiview predictions into an SVBRDF texture atlas.

Our study is motivated by recent work on object texturing using 2D diffusion models \cite{Richardson2023TEXTure,chen2023text2tex}. Combining these methods with single-image SVBRDF estimation, we design a pipeline that takes as input the geometry of a 3D scene and outputs an SVBRDF texture atlas for that scene, which we use to evaluate the various design choices involved. This pipeline is controllable, leveraging conditional image generation \cite{zhang2023adding} and image re-projection and inpainting \cite{Avrahami_2022_CVPR} to iteratively generate multiple views of the input scene conditioned on its geometry and user-provided text or image prompts. Complementing multiview image generation with per-view material estimation allows us to obtain SVBRDF maps that we project onto a texture atlas to render the scene from arbitrary viewpoints and under arbitrary lighting.

The main contribution of our work resides in the exploration of various design choices for this material generation pipeline. Specifically, we compare state-of-the-art SVBRDF estimation methods in terms of both single-image accuracy and multiview coherence. Furthermore, we assess whether SVBRDF prediction improves when performed on deep features generated by the diffusion model \cite{luo2023dhf} rather than on its RGB output. We also study the benefit of providing geometry information (depth and normal buffers) to the material estimation module. Surprisingly, this analysis reveals that a simple UNet that regresses SVBRDF maps is competitive with more complex alternatives.

In summary, this paper introduces two contributions:
\begin{itemize}
    \item An analysis of the design space of single-image SVBRDF estimation for multiview material generation.
    \item Based on this analysis, a fast and controllable pipeline to generate SVBRDF textures over indoor scenes, combining generative image models and SVBRDF predictors.
\end{itemize}

Code, model weights and data have been released at: \href{https://github.com/graphdeco-inria/svbrdf-evaluation}{github.com/graphdeco-inria/svbrdf-evaluation}

\section{Related Work}
Our work leverages recent diffusion models for semi-automated visual content creation. We refer readers to the recent tutorial by Mitra et al. \cite{Mitra2024} and the report by Po et al. \cite{STARDiffusion} for an overview of this very active field, and focus our discussion on the methods most related to our goal of generating SVBRDF textures over 3D scenes.

\paragraph*{3D asset texturing.}
Automatic texturing of existing 3D assets has seen rapid progress with the introduction of high quality image generation methods~\cite{Rombach_2022_CVPR, ramesh2022hierarchical}. Two main directions have been explored, based on either Score Distillation Sampling (SDS)~\cite{poole2023dreamfusion} or on multiview texturing. SDS-based methods rely on the gradient provided by an image generative model to optimize a 3D representation of an object or scene geometry and/or texture \cite{chen2024scenetex, deng2024flashtex}. Early versions of these methods rely on many diffusion steps of the generative model, often requiring dozens of minutes, while the more recent FlashTex \cite{deng2024flashtex} proposed a hybrid approach leveraging strong initialization and neural hashgrids for faster optimization, requiring only 6 minutes for texturing an object. Still, SDS-based approaches remain slower than multiview texturing methods~\cite{zeng2024paint3d, Richardson2023TEXTure, chen2023text2tex, TexPainter2024Zhang, ceylan2024matatlas, perla2024easitex, zhang2025instex, wang2024boosting, liu2024text} that generate views around an object conditioned on its geometry, and project these images over the object to form a texture atlas. Such methods have so far been mostly demonstrated on isolated objects to create RGB textures. We build on these methods to design a pipeline that works on 3D scenes, and we augment RGB image generation with material prediction to produce SVBRDF atlases.

Several approaches also extended Score Distillation Sampling \cite{DreamMat2024Zhang,youwang2024paintit,TextureDreamer2024CVPR} and multiview texturing \cite{zhang2024mapa, ceylan2024matatlas, fang2024makeitreal} to create materials over 3D objects. The former family of methods relies on expensive inverse rendering to optimize material parameters such that they produce images with a target appearance. In contrast, the latter family relies on retrieval in a library of procedural materials to assign SVBRDFs to segments in the generated RGB images. We share the same goal of augmenting generated images with material information, but we evaluate the performance of SVBRDF prediction rather than material retrieval in the context of multiview texturing. Furthermore, the above methods target isolated objects while we target extended indoor scenes that exhibit significant occlusions across views. Recent concurrent work \cite{vainer2024jointly, huang2024materialanything} generate multi-view material maps but also target isolated objects.

Finally, several methods have been proposed to automatically generate 3D objects or scenes, including their geometry and textures, from a single image or text prompt. Such methods rely on SDS-based optimization of shape and colors \cite{poole2023dreamfusion,Lin_2023_CVPR}, on retrieval from datasets of 3D objects and materials~\cite{Yan:2023:PSDR-Room}, on iterative generation of multiview-consistent images/depth/material maps \cite{Tang2023mvdiffusion,hoellein2023text2room,Liu2023Zero1to3ZO, zhang2024clay}, or on volumetric generation \cite{hong2023lrm, wei2024meshlrm}. While we focus on the different scenario of applying materials on an existing scene, we share their motivation for providing high-level controls on content creation, and some of their strategies for iterative generation of multiview-consistent images \cite{hoellein2023text2room}.

\begin{figure*}[ht]
  \centering
  \includegraphics[width=\textwidth]{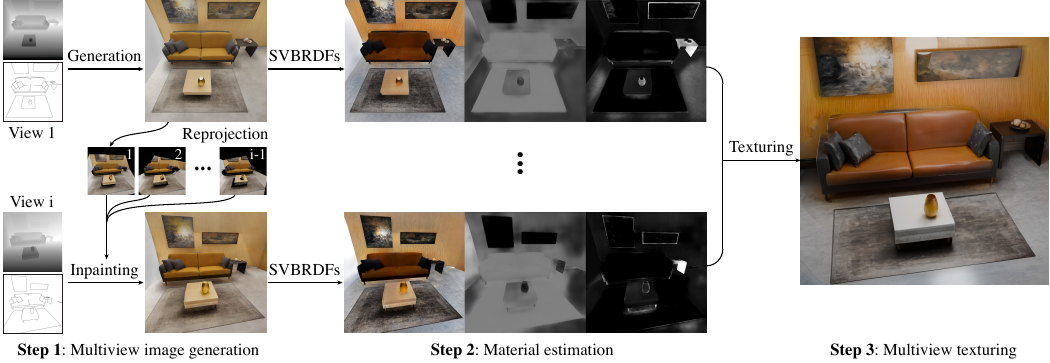}
  \caption{\textbf{Overview of the SVBRDF texturing pipeline.} 
  We first generate a sequence of views of the scene using an image diffusion model conditioned on depth and contours (Step 1). While the first view is generated entirely (top left), subsequent views are filled in by re-projecting existing views and inpainting holes revealed by disocclusions (bottom left). We then estimate SVBRDF maps for each view (Step 2), and merge them to form a texture atlas for the scene, enabling physically-based rendering with diverse materials (Step 3).
  }
  \label{fig:overview}
\end{figure*}

\paragraph*{Predicting SVBRDFs.}
Predicting reflectance of surfaces from photograph(s) has been a long-standing challenge in Computer Graphics and Vision. When many (consistent) views are available, inverse rendering aims at reconstructing material properties through (gradient-based) optimization~\cite{Mitsuba3, Li:2018:DMC, Ramamoorthi2001}. Alternatively, deep neural networks have been used to predict material parameters for acquisition with limited signal, or when the signal is not guaranteed to be consistent across views -- as is our case with generated images. Various neural methods were proposed for acquisition on flat surfaces~\cite{DADDB18, DADDB19, GuoMaterialGAN2020, vecchio2023controlmat}, objects~\cite{vainer2024collaborative, li2018learning, deschaintre21, hong2024supermat} and, more recently, entire scenes from a single view~\cite{li2020inverse,kocsis2024iid,RGB2X2024Zeng, wang2025materialist} or multiple views \cite{choi2023mair, DiffusionRenderer, he2024neural}. We focus on single-image methods due to the lack of a public dataset with paired renderings, SVBRDF maps, and cameras from multiple views.  The decomposition of scene reflectance and geometry has received significant attention in recent years, leveraging the progress of generative models through fine-tuning~\cite{kocsis2024iid, RGB2X2024Zeng} or through manipulation and analysis of their internal features ~\cite{chen2023beyond,bhattad2023stylegan,zhao2023unleashing,luo2024readoutguidance}. A key aspect of our work is to evaluate these different approaches in the context of material design for 3D scenes and to compare them in terms of decomposition accuracy and consistency between views.

\section{A Multiview Pipeline for Fast Appearance Modeling of 3D Scenes}
\label{sec:pipeline}
We first describe a fast pipeline allowing users to apply materials over the 3D geometry of an untextured indoor scene by specifying high-level design goals in the form of text prompts or example images. This pipeline will allow us to investigate the design choices required for this task.

We illustrate this pipeline in Figure~\ref{fig:overview}. Most of its components operate in image space, benefiting from the complementary strengths of diffusion models for controllable generation of multiple views of the scene, SVBRDF predictors for material estimation within each view, and multiview reprojection for texturing the scene with the obtained SVBRDF maps. 
Importantly, each operation is fast to compute, allowing to generate materials over a scene in minutes.
We describe the first and third steps of this pipeline -- multiview image generation and multiview texturing -- before discussing and evaluating important design choices for SVBRDF prediction in Section~\ref{sec:mat_extraction}.

\subsection{Multiview Image Generation}
\begin{figure}[!t]
  \centering
  \includegraphics[width=0.47\textwidth]{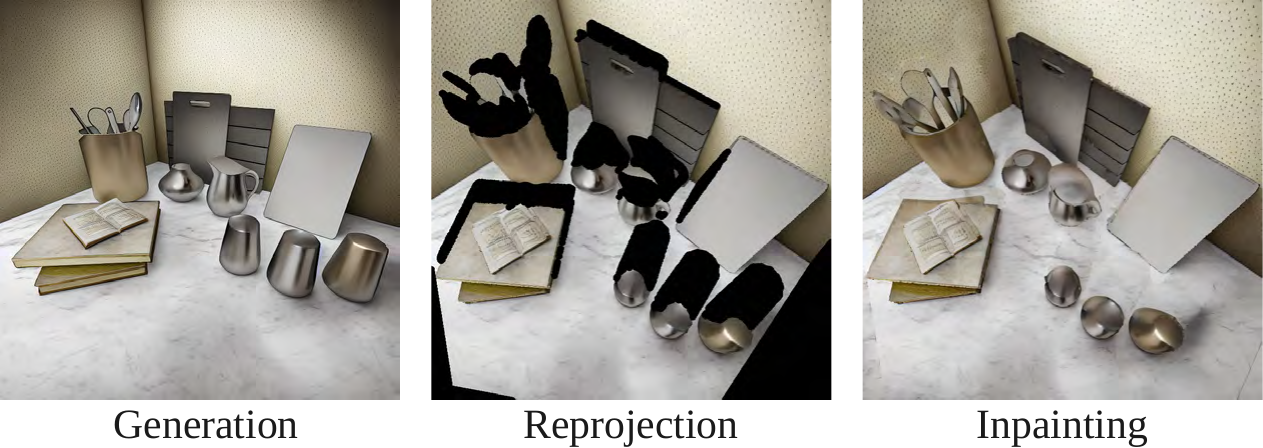}
  \caption{We progressively generate multiple coherent views of the scene by iterating over image generation conditioned on geometry, image reprojection into a new viewpoint, and image inpainting.}
  \label{fig:inpainting}
\end{figure}

The pipeline that underpins our study takes as input the geometry of an indoor 3D scene and a sequence of viewpoints that cover that scene well. 
For all the results in the paper, we use five viewpoints. We assume the first view is provided by the user. We generate the other four views by an offset of 25 degrees in latitude and longitude.
The first step consists in generating multiview-consistent images of that scene observed from the given viewpoints. Inspired by previous work on generative single-object texturing and novel view synthesis \cite{Richardson2023TEXTure,chen2023text2tex, perla2024easitex,invs2023,hoellein2023text2room}, we create consistent images of the scene by alternating image generation, image reprojection, and image inpainting (Figure~\ref{fig:inpainting}). 

\paragraph*{Image generation.}
We initiate the process with the first viewpoint, from which we render depth and contour maps that we use to condition an image generative model -- ControlNet with a Stable Diffusion backbone \cite{zhang2023adding} -- to obtain a photorealistic image aligned with the scene geometry. We optionally allow users to further control the content of the generated image by providing a text prompt, or an example image that contains representative materials to be generated. In the case where an example image is provided, we use the IP-Adapter approach \cite{ye2023ipadapter} to obtain a text embedding for that image, which can then be used as a condition for Stable Diffusion.
We combine all these conditions using the Multi-ControlNet pipeline from the Diffuser library \cite{vonplatenetal2022diffusers}. 

We next project the generated image from the first viewpoint into the second one using their respective cameras and depth maps, which ensures that the portions of the scene observed in both viewpoints are consistent. 
We hypothesize that while image reprojection only provides approximate placement of highlights, this approximation provides sufficient visual cues to estimate intrinsic appearance maps (metallic, roughness, etc.) because these quantities depend more on the sharpness and contrast of highlights than on their precise position.
Our experiments demonstrate that, while potentially counter-intuitive, the combination of simple reprojection and per-view SVBRDF estimation can suffice to produce coherent SVBRDFs, which can in turn be aggregated into a single texture atlas for physically-based rendering of specular and glossy reflections.

However, reprojection leaves holes at disocclusions, especially in cluttered indoor scenes where visibility changes significantly between viewpoints. 
We fill in these holes using image inpainting, as described in the next paragraph. We repeat the reprojection and inpainting steps to progressively populate all viewpoints, using all images generated so far to create the image of the next viewpoint.

\paragraph*{Image inpainting.}
Reprojection provides us with a partially-covered image in the new viewpoint, as well as an occlusion mask indicating the parts to be filled-in. We feed this information to a ControlNet trained to perform inpainting \cite{controlnetv11}, along with the same text prompt used to generate the first image, if any. We also provide a text embedding of the first generated image, which helps in generating similar content in inpainted regions.
We further improve alignment of the inpainted image to the underlying scene geometry by also conditioning ControlNet with depth and contour maps, as done when generating the image of the first viewpoint. 
In our experiments, the holes to inpaint cover up to 25\% of the image.

\subsection{Multiview Texturing}
The above iterative procedure generates one image for each input viewpoint while ensuring that the parts seen from several viewpoints are consistent. Feeding each such image to a SVBRDF predictor gives material parameter maps for all viewpoints, as described in the next section. The last step of our pipeline consists in merging this image-space information into a common, scene-space texture atlas. 

In contrast to related work on object texturing that populates the texture atlas incrementally \cite{Richardson2023TEXTure,chen2023text2tex, perla2024easitex}, we take inspiration from photogrammetry where all photographs are merged at once to select the best observations available for each texel. While several algorithms exist to perform this task, we use the one implemented in MeshLab \cite{CCCS08} for its simplicity and speed. This method takes as input the 3D mesh of the scene, its texture coordinates, and the multiple images of the scene with their respective camera parameters. It then assigns values to each texel by blending its observations according to geometric and color criteria. Since the SVBRDF maps we want to merge contain different quantities (albedo, roughness, metallic), we apply the method on each quantity independently.

\section{Evaluating SVBRDF Predictors for Multiview Appearance Modeling} \label{sec:mat_extraction}

The above pipeline for generative material modeling requires extracting material maps from multiple images created via conditional image diffusion. 
This material estimation needs to be both accurate and consistent to facilitate the subsequent texture merging process. We now describe the various design choices that emerge from this setup, and analyze the impact of these choices.

\subsection{A Design Space of SVBRDF Predictors}

\paragraph*{Scope of the study.}
The pipeline we have described in the previous section generates multiple photorealistic images of the 3D scene. A first solution that comes to mind to estimate the materials in that scene is physically-based inverse rendering, i.e. optimizing for material parameters that best reproduce the multiple images \cite{azinovic2019inverse, nimierdavid2021material, fipt2023}. However, this solution is not practical because even though the pipeline generates plausible images, 
the lighting conditions are unknown, and might even be inconsistent due to the simple reprojection employed to iteratively generate the views, preventing the use of physically-based inverse rendering. 

\begin{figure*}[ht]
  \centering
  \includegraphics[width=\textwidth]{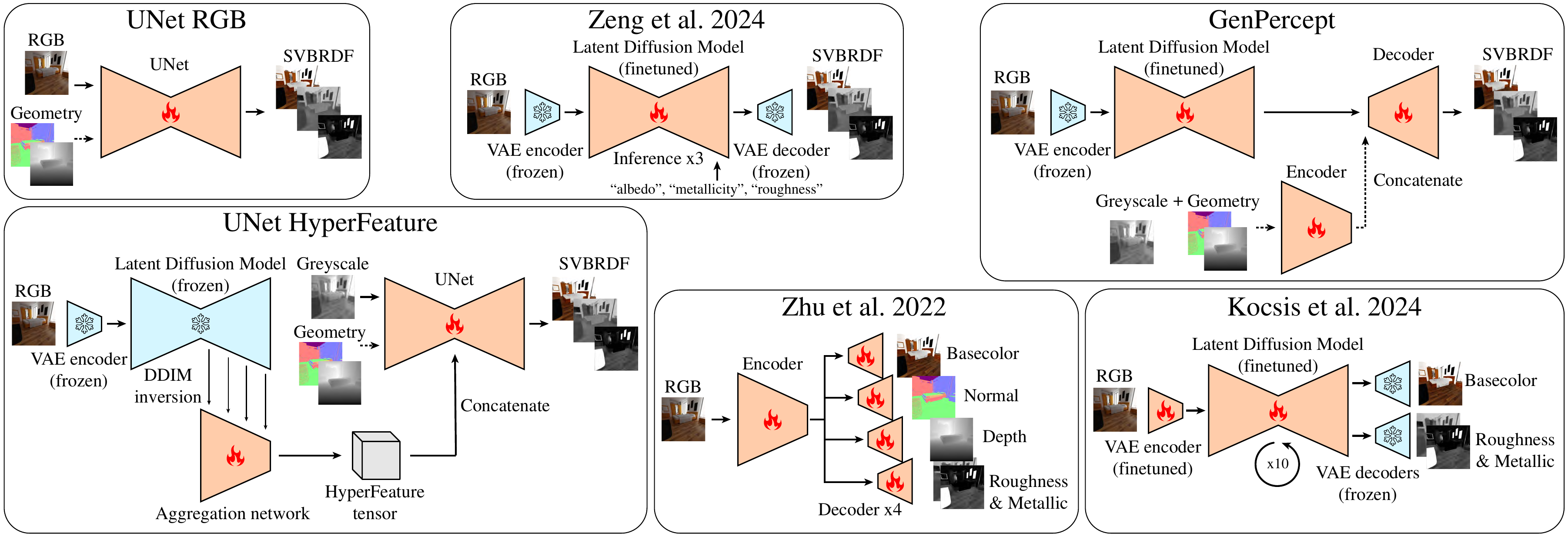}
  \caption{\textbf{SVBRDF predictors.} We evaluate several architectures for SVBRDF predictions, taking as input an RGB image and optional geometry buffers (dotted lines).
  }
  \label{fig:design_space}
\end{figure*}

These considerations motivate our choice of focusing our study on single-view SVBRDF prediction methods, and assessing whether these methods produce material parameters of sufficient quality and coherence to be combined into a single texture atlas. Specifically, we conducted experiments on two key dimensions of the design space of SVBRDF predictors: the choice of neural architecture, and the choice of data channels provided as input to that architecture, as summarized in Figure~\ref{fig:design_space}. 
For each of these choices, we evaluate the per-view accuracy of the predicted SVBRDFs, as well as their coherence across views.

\paragraph*{Choice of architecture.}
Several neural network architectures have been proposed over the past few years to predict SVBRDF maps from a single image of an indoor scene.
Early work relies on convolutional neural networks to predict material maps from RGB images, trained on paired datasets of images and ground-truth material maps to minimize regression (MAE, MSE) and perceptual losses (VGG, LPIPS).
We evaluate Zhu et al.'s MGNet \cite{zhu2022learning} as a representative architecture of this family, using their model pre-trained on their InteriorVerse dataset, which we use for all comparisons. We also evaluate the performance of a standard UNet trained on the same dataset, which we call the \textbf{UNet-RGB} architecture.

More recently, generative models, and in particular diffusion models, have been repurposed for SVBRDF estimation. A key difference with regression-based methods is that generative models treat the input RGB image as guidance within a stochastic generation process that seeks to capture the distribution of SVBRDF maps that best explain the image. As a result, these methods can output multiple plausible interpretations of a given input.
We evaluate the methods of Kocsis et al.~\cite{kocsis2024iid} and Zeng et al.~\cite{RGB2X2024Zeng} as representative works in this category, using their model pre-trained on InteriorVerse. We follow the recommendations of Kocsis et al. and average 10 predictions of their model to form their output, unless specified otherwise.
Finally, we also evaluate the single-step approach \textbf{GenPercept} by Xu et al. \cite{xu2024genpercept}, which repurposes a diffusion model to perform dense perception tasks deterministically. We trained their method on InteriorVerse for our evaluation.

\begin{figure*}[!h]
    \centering
    \includegraphics[width=0.9\textwidth]{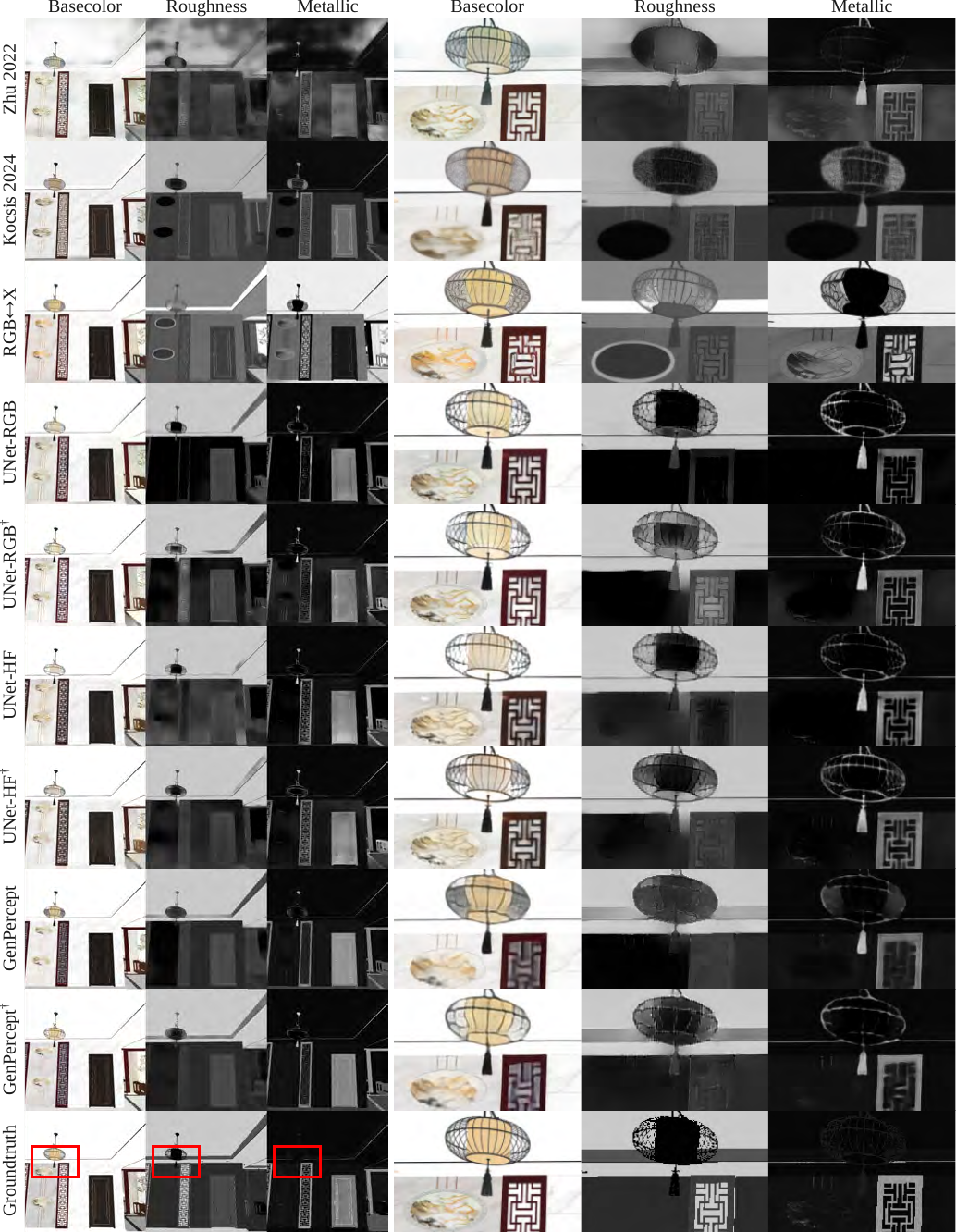}
    \caption{Qualitative results of SVBRDF estimation for a test set image from Interiorverse, with closeups on the right. 
    $^{\dag}$Indicates training without normals and depth maps as input. Note how different methods provide a different tradeoff between accuracy, sharpness, and fine details.}
    \label{fig:qualitative_svbrdf}
\end{figure*}

\paragraph*{Choice of input channels.}
The architectures discussed above were developed to take photographs only as input. Yet, our target application -- generative material design for 3D scenes -- provides us with additional information, such as the scene geometry, that can help the SVBRDF predictor. We have experimented with several choices of additional information for the UNet and the GenPercept architectures that we retrain on InteriorVerse.

First, we can complement the RGB input with depth and normal maps to provide the neural networks with geometric information about the 3D scene, which we expect to help in recovering sharper maps along surface discontinuities, and in distinguishing shading from albedo variations. 

Second, since the RGB images we take as input are the result of a generative process, we can complement each image with deep activation features extracted from the neural network that generated that image. Indeed, recent studies suggest that the feature maps produced by generative models of images carry semantic information about the content of the images they generate, allowing the recovery of semantic labels \cite{baranchuk2021labelefficient,zhang2021datasetgan}, keypoint correspondences \cite{luo2023dhf,pan2023_DragGAN,hedlin2023unsupervised}, depth, normal and albedo \cite{chen2023beyond,bhattad2023stylegan,zhao2023unleashing}. 
To evaluate whether such deep features benefit SVBRDF prediction, we follow the approach of Luo et al. \cite{luo2024readoutguidance,luo2023dhf} to extract so-called \emph{hyperfeatures} from the denoising UNet of Stable Diffusion used to generate the first view or of the one used to inpaint the other views. We then concatenate these hyperfeatures with the feature maps of the bottleneck of the SVBRDF prediction UNet. We call this the \textbf{UNet-HF} (Hyperfeatures) architecture. 
While the Stable Diffusion activation features (which aggregate into a hyperfeatures tensor) are readily available for the images generated by the pipeline, they are not provided for the images of the InteriorVerse dataset we use for training. We solve this by performing inverse DDIM sampling \cite{song2021denoising} to ``invert'' each training image using Stable Diffusion, providing us with the UNet activation features corresponding to that image.

\paragraph*{Implementation details.}
We base our implementation of the \textbf{UNet-RGB} architecture on the denoising diffusion UNet from the Diffusers library \cite{vonplatenetal2022diffusers}, and our implementation of the \textbf{UNet-HF} architecture on the Readout Guidance \cite{luo2024readoutguidance} codebase for extracting hyperfeatures from Stable Diffusion's denoising UNet, using 11 timesteps and a projection dimension of 384 (as in~\cite{luo2023dhf}). 
For single-step diffusion, we follow the implementation of \textbf{GenPercept} provided by Xu et al. \cite{xu2024genpercept}.
We provide additional details for each architecture in supplementary.

\begin{table*}[!h]
    \centering
    \resizebox{\linewidth}{!}{
    \begin{tabular}{l|ccc|ccc|ccc|}
    \cline{2-10}
    & \multicolumn{3}{c|}{\textsc{Basecolor}} & \multicolumn{3}{c|}{\textsc{Roughness}} & \multicolumn{3}{c|}{\textsc{Metallic}} \\
    \multicolumn{1}{c|}{}                                               & PSNR $\uparrow$   & SSIM $\uparrow$   & LPIPS $\downarrow$    & PSNR $\uparrow$   & SSIM $\uparrow$   & LPIPS $\downarrow$    & PSNR $\uparrow$   & SSIM $\uparrow$   & LPIPS $\downarrow$ \\ \hline
    \multicolumn{1}{|l|}{Zhu2022 \cite{zhu2022learning}}                & \cellcolor[HTML]{E0A4A4}17.11          & \cellcolor[HTML]{E0A4A4}0.668          & \cellcolor[HTML]{E0A4A4}0.229          & \cellcolor[HTML]{EABFAF}13.38          & \cellcolor[HTML]{E9BCAE}0.348          & \cellcolor[HTML]{E0A4A4}0.422          & \cellcolor[HTML]{F3D6B8}14.93          & \cellcolor[HTML]{F9E6BF}0.676          & \cellcolor[HTML]{F2D2B7}0.419          \\
    \multicolumn{1}{|l|}{Kocsis2024 \cite{kocsis2024iid}}               & \cellcolor[HTML]{F7E0BC}19.83          & \cellcolor[HTML]{F5DCBB}0.743          & \cellcolor[HTML]{F5DBBB}0.163          & \cellcolor[HTML]{F2D4B7}14.83          & \cellcolor[HTML]{F2D3B7}0.469          & \cellcolor[HTML]{F1D0B6}0.337          & \cellcolor[HTML]{FAE9C0}17.95          & \cellcolor[HTML]{FEF2C4}0.805          & \cellcolor[HTML]{FBE9C0}0.327          \\
    \multicolumn{1}{|l|}{RGB$\leftrightarrow$X \cite{RGB2X2024Zeng}}    & \cellcolor[HTML]{FBECC1}20.35          & \cellcolor[HTML]{FBEBC1}0.763          & \cellcolor[HTML]{FFF5C5}0.131          & \cellcolor[HTML]{E0A4A4}11.38          & \cellcolor[HTML]{E0A4A4}0.220          & \cellcolor[HTML]{E5AFA9}0.401          & \cellcolor[HTML]{E0A4A4}6.68           & \cellcolor[HTML]{E0A4A4}-0.019         & \cellcolor[HTML]{E0A4A4}0.604          \\
    \hline \hline
    \multicolumn{1}{|l|}{UNet-RGB$^{\dag}$}                             & \cellcolor[HTML]{FAEAC0}20.26          & \cellcolor[HTML]{FEF2C4}0.772          & \cellcolor[HTML]{ECEBBA}0.126          & \cellcolor[HTML]{FDF1C3}16.92          & \cellcolor[HTML]{FFF5C5}0.642          & \cellcolor[HTML]{FFF5C5}0.265          & \cellcolor[HTML]{FEF3C4}19.52          & \cellcolor[HTML]{EEEDBB}0.834          & \cellcolor[HTML]{F7F1C0}0.274          \\
    \multicolumn{1}{|l|}{UNet-RGB}                                      & \cellcolor[HTML]{DBE3B0}21.66          & \cellcolor[HTML]{A8C992}\textbf{0.803} & \cellcolor[HTML]{AFCC96}0.109          & \cellcolor[HTML]{ECECBA}17.73          & \cellcolor[HTML]{D3DFAC}0.672          & \cellcolor[HTML]{DEE4B1}0.241          & \cellcolor[HTML]{E8EAB8}20.13          & \cellcolor[HTML]{C7D9A4}0.843          & \cellcolor[HTML]{AFCD96}0.256          \\
    \multicolumn{1}{|l|}{UNet-HF$^{\dag}$}                              & \cellcolor[HTML]{FFF5C5}20.74          & \cellcolor[HTML]{FFF5C5}0.775          & \cellcolor[HTML]{FDEFC3}0.139          & \cellcolor[HTML]{FFF5C5}17.20          & \cellcolor[HTML]{FFF5C5}0.642          & \cellcolor[HTML]{FFF3C5}0.269          & \cellcolor[HTML]{FFF5C5}19.82          & \cellcolor[HTML]{FFF5C5}0.830          & \cellcolor[HTML]{FFF5C5}0.280          \\
    \multicolumn{1}{|l|}{UNet-HF}                                       & \cellcolor[HTML]{EBEBBA}21.25          & \cellcolor[HTML]{E4E7B5}0.784          & \cellcolor[HTML]{FFF3C4}0.134          & \cellcolor[HTML]{FDF4C4}17.27          & \cellcolor[HTML]{FDF1C3}0.626          & \cellcolor[HTML]{FFF5C5}0.265          & \cellcolor[HTML]{F3EFBE}19.99          & \cellcolor[HTML]{FFF5C5}0.830          & \cellcolor[HTML]{FFF5C5}0.276          \\
    \multicolumn{1}{|l|}{GenPercept$^{\dag}$}                           & \cellcolor[HTML]{A8C992}\textbf{22.94} & \cellcolor[HTML]{B5D09A}0.799          & \cellcolor[HTML]{A8C992}\textbf{0.107} & \cellcolor[HTML]{A8C992}\textbf{19.57} & \cellcolor[HTML]{A8C992}\textbf{0.701} & \cellcolor[HTML]{A8C992}\textbf{0.201} & \cellcolor[HTML]{A8C992}\textbf{20.97} & \cellcolor[HTML]{A8C992}\textbf{0.850} & \cellcolor[HTML]{AFCD96}0.256          \\
    \multicolumn{1}{|l|}{GenPercept}                                    & \cellcolor[HTML]{B3CF99}22.67          & \cellcolor[HTML]{C8D9A5}0.793          & \cellcolor[HTML]{BDD49E}0.113          & \cellcolor[HTML]{B4CF99}19.26          & \cellcolor[HTML]{BAD29D}0.689          & \cellcolor[HTML]{A9C992}0.202          & \cellcolor[HTML]{B3CF99}20.83          & \cellcolor[HTML]{B6D09A}0.847          & \cellcolor[HTML]{A8C992}\textbf{0.254} \\ \hline
    \end{tabular}
    }
    \caption{\textbf{Quantitative analysis of SVBRDF estimation}. The metrics are computed over the InteriorVerse test dataset (2633 images),
    color coded between \BGcolor{E0A4A4}{w}\BGcolor{E4B0A9}{o}\BGcolor{E9BCAE}{r}\BGcolor{EEC9B3}{s}\BGcolor{F3D5B8}{t}\BGcolor{F7E2BD}{ }\BGcolor{FCEEC2}{a}\BGcolor{F9F2C2}{n}\BGcolor{EBEBBA}{d}\BGcolor{DEE5B2}{ }\BGcolor{D1DEAA}{b}\BGcolor{C3D7A2}{e}\BGcolor{B6D09A}{s}\BGcolor{A8C992}{t}. 
    $^{\dag}$Indicates training without normals and depth maps as input.}
    \label{table:quantitative}
\end{table*}

\subsection{Accuracy and Coherence of SVBRDF Predictors}

We investigate two main dimensions in the design space of SVBRDF generations for 3D scenes -- the choice of \emph{neural architecture} and the effect of different \emph{inputs} for SVBRDF estimation. For each dimension, we evaluate the accuracy of SVBRDF predictions for single images as well as the coherence of the predictions over multiple views. We provide both quantitative and qualitative evaluation of both criteria.

\paragraph*{Training data and procedure.}
We use the pre-trained models provided by Zhu et al. \cite{zhu2022learning}, Kocsis et al.~\cite{kocsis2024iid} and Zeng et al.~\cite{RGB2X2024Zeng}, which have all been trained on the InteriorVerse dataset~\cite{zhu2022learning}.
In addition, we also train the UNet models and the single-step diffusion model~\cite{xu2024genpercept} on InteriorVerse. 

We conduct our evaluation on the InteriorVerse test dataset because it is the only public dataset containing the BRDF maps we are interested in. 
The dataset provides viewspace BRDF maps for basecolor (called "albedo"), metallic and roughness. The basecolor is modulated by the metallic, and encodes both diffuse albedo and specular color, as commonly used in physically-based pipelines \cite{burley2012physically}.
As noted in RGB$\leftrightarrow$X~\cite{RGB2X2024Zeng} the rendering images of InteriorVerse contain Monte-Carlo noise, which we denoise using Mitsuba's Optix integration. We crop each image to get a square which we resize to $512^{2}$ during training. 
Additionally, some images of the InteriorVerse dataset contain very few objects or very little detail, and are sometimes rendered from cameras pointing outside the scene. We remove such data (4\% of the dataset) by removing files below 2MB. 
We will release this curated dataset upon publication, together with the code and network weights.

\begin{figure}[!t]
    \centering
    \includegraphics[width=0.47\textwidth]{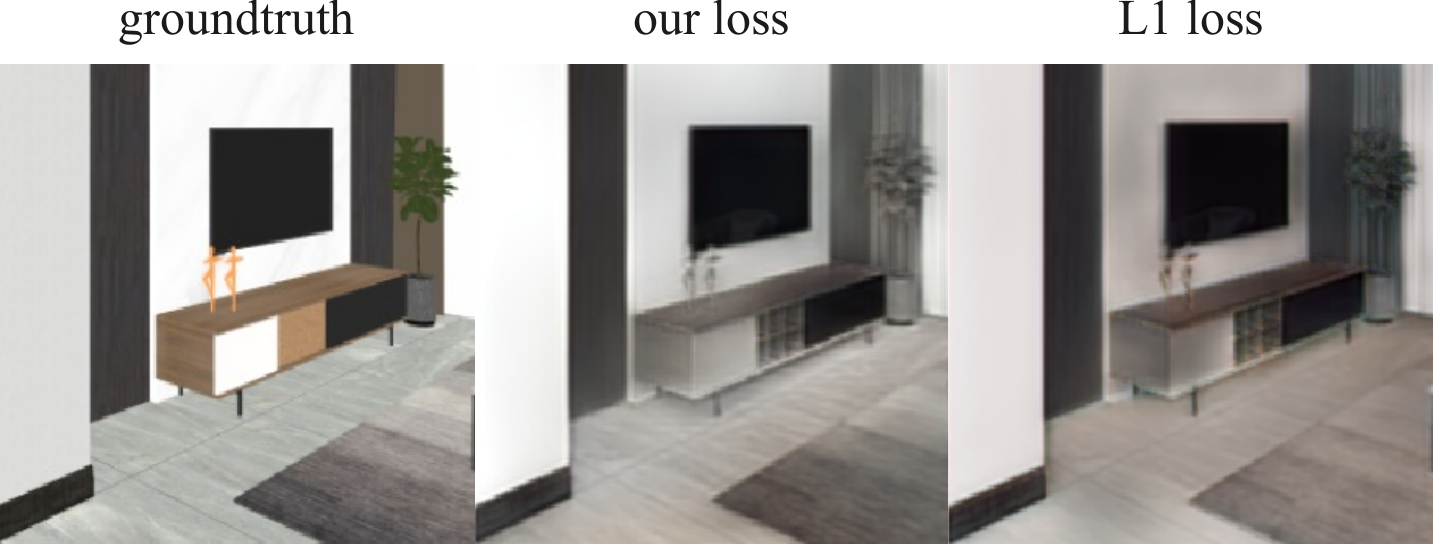}
    \caption{
    When training using the FLIP loss colorspace \cite{Andersson2020FLIP}, the basecolor converges faster to the right hue and luminance than when using a simple L1 loss in RGB colorspace. Note in particular the hue of the wall and floor. The results are shown after only a few epochs of training to emphasize differences.
    \label{fig:fliploss}
    }
\end{figure}

Regarding the training procedure of these predictors, we use a combination of L-PIPS loss~\cite{zhang2018perceptual} $\mathcal{L}^{vgg}$ and L1 loss computed over the perceptual color space proposed by \FLIP \cite{Andersson2020FLIP} $\mathcal{L}_{1}^{FLIP}$ for the basecolor, which we found to behave better than the L1 loss in RGB space (see Fig.~\ref{fig:fliploss}).
We use a regular L1 loss for both metallic and roughness.
The training loss is thus the sum of three groups of terms corresponding to basecolor, metallic and roughness, respectively weighted by $\alpha_b$, $\alpha_m$, $\alpha_r$:
\begin{equation}
    \alpha_b ( \mathcal{L}_{1}^{FLIP} + \lambda_{b} \cdot \mathcal{L}^{vgg} ) + \alpha_m \cdot \mathcal{L}_{1} + \alpha_r ( \mathcal{L}_{1} + \lambda_{r} \cdot \mathcal{L}^{vgg} ),
\end{equation}
where $\alpha_b=1.0$, $\alpha_m=2.0$, $\alpha_r=0.5$, and $\lambda_{b}=\lambda_{r}=0.5$, which we empirically find to work best.

We train each estimator with a maximum budget of 8 days using 4 GPUs (either RTX 6000s or RTX 8000s). We train all networks with an Adam optimizer and a learning rate of 10e-5.

\paragraph*{Single-view prediction.}
We first study how the different architectures and inputs perform on a single-image prediction task.

We run each method on the InteriorVerse test set, composed of 2633 HDR images, which we process with the tone-mapping and normalization algorithms recommended by the authors of each method.
As suggested by \cite{kocsis2024iid}, we lift the luminance ambiguity in the estimation by reporting scale-invariant metrics for the basecolor. This is computed by normalizing the output of each method by the ratio between the ground-truth and the predicted per-channel means. We report standard quantitative error metrics (PSNR, SSIM and LPIPS) in Tab.~\ref{table:quantitative} and we provide representative qualitative comparisons in Fig.~\ref{fig:qualitative_svbrdf}. We provide additional qualitative results in supplemental materials.

The simple UNet-RGB with geometry cues performs surprisingly well on all metrics, being only outperformed by GenPercept (irrespectively of the use of additional geometric information). The UNet-HF with geometry cues is placed third.
From visual inspection (Fig.~\ref{fig:qualitative_svbrdf}), we see that the method of Kocsis et al.~\cite{kocsis2024iid} and RGB$\leftrightarrow$X~\cite{RGB2X2024Zeng} create smooth, piecewise-constant maps, as is the case for ground truth SVBRDFs. However, these piecewise constant maps are not always accurate, in particular for roughness and metallic maps where different objects made of the same materials are assigned different -- often erroneous -- values, as is the case for the insets in the corresponding rows of Fig.~\ref{fig:qualitative_svbrdf} where the roughness and metallic maps should be uniform over the wall rather than being correlated with the basecolor. MGNet~\cite{zhu2022learning} tends to produce splotchy results in all maps, compromising quality. Finally, GenPercept \cite{xu2024genpercept} is the most accurate according to numerical metrics (Tab.~\ref{table:quantitative}), yet struggles to recover fine details, as is the case of the thin structures of the lamp and wall panel in Fig.~\ref{fig:qualitative_svbrdf}.
Overall, each method provides different tradeoffs between sharpness and accuracy, as illustrated by additional examples in the supplemental material. The use of additional geometric inputs tends to improve the overall accuracy of the predictors.

\begin{table}[t]
    \centering
    \resizebox{\linewidth}{!}{
    \begin{tabular}{l|cccc}
        \cline{2-4}
                                                                    & \multicolumn{3}{c|}{Consistency $\downarrow$}                                              &         \\ \hline
        \multicolumn{1}{|l|}{Methods}                               & \multicolumn{1}{c}{Basecolor}            & \multicolumn{1}{c}{Roughness}        & \multicolumn{1}{c|}{Metallic} &         \multicolumn{1}{c|}{Runtime} \\ \hline
        \multicolumn{1}{|l|}{ \cite{kocsis2024iid}$ ^{\times1} $}   & \cellcolor[HTML]{E0A4A4}0.281          & \cellcolor[HTML]{E4AFA9}0.227         & \multicolumn{1}{c|}{\cellcolor[HTML]{F4D7B9}0.219}            & \multicolumn{1}{c|}{ $ \simeq 6 $ s  } \\ 
        \multicolumn{1}{|l|}{ \cite{kocsis2024iid}$ ^{\times10} $}  & \cellcolor[HTML]{F6DDBB}0.124          & \cellcolor[HTML]{F8ECC0}0.100         & \multicolumn{1}{c|}{\cellcolor[HTML]{FBEAC1}0.104}            & \multicolumn{1}{c|}{ $ 60.8 \pm 1.3$ s } \\ 
        \multicolumn{1}{|l|}{\cite{RGB2X2024Zeng}}                  & \cellcolor[HTML]{F2D2B7}0.155          & \cellcolor[HTML]{E0A4A4}0.247         & \multicolumn{1}{c|}{\cellcolor[HTML]{E0A4A4}0.542}            & \multicolumn{1}{c|}{ $ 11.1 \pm 0.2$ s } \\ \hline \hline
        \multicolumn{1}{|l|}{\cite{zhu2022learning}}                & \cellcolor[HTML]{F7EBBF}0.073          & \cellcolor[HTML]{F8ECC0}0.100         & \multicolumn{1}{c|}{\cellcolor[HTML]{FBE9C0}0.105}            & \multicolumn{1}{c|}{ $ 41 \pm 5$ ms } \\ 
        \multicolumn{1}{|l|}{UNet-RGB}                              & \cellcolor[HTML]{CDD9A7}0.064          & \cellcolor[HTML]{D5DCAB}0.078         & \multicolumn{1}{c|}{\cellcolor[HTML]{D9DEAE}0.060}             & \multicolumn{1}{c|}{$ 36 \pm 1$ ms} \\ 
        \multicolumn{1}{|l|}{UNet-RGB$^{\dag}$}                     & \cellcolor[HTML]{FAE9C0}0.089          & \cellcolor[HTML]{FAE7BF}0.117         & \multicolumn{1}{c|}{\cellcolor[HTML]{FCEEC2}0.077}            & \multicolumn{1}{c|}{$ 36 \pm 2$ ms} \\ 
        \multicolumn{1}{|l|}{UNet-HF}                               & \cellcolor[HTML]{A8C992}\textbf{0.056} & \cellcolor[HTML]{A8C992}\textbf{0.050}& \multicolumn{1}{c|}{\cellcolor[HTML]{A8C992}\textbf{0.041}}   & \multicolumn{1}{c|}{$ 335 \pm 34$ ms} \\ 
        \multicolumn{1}{|l|}{UNet-HF$^{\dag}$}                      & \cellcolor[HTML]{BAD19C}0.060          & \cellcolor[HTML]{BED29E}0.064         & \multicolumn{1}{c|}{\cellcolor[HTML]{BAD19C}0.048}            & \multicolumn{1}{c|}{$ 409 \pm 12$ ms} \\ 
        \multicolumn{1}{|l|}{GenPercept}                            & \cellcolor[HTML]{F2E9BC}0.072          & \cellcolor[HTML]{FCEDC2}0.104         & \multicolumn{1}{c|}{\cellcolor[HTML]{F1E9BB}0.069}            & \multicolumn{1}{c|}{$ 69 \pm 4$ ms} \\ 
        \multicolumn{1}{|l|}{GenPercept$^{\dag}$}                   & \cellcolor[HTML]{FCEEC2}0.075          & \cellcolor[HTML]{FBEBC1}0.108         & \multicolumn{1}{c|}{\cellcolor[HTML]{F1E9BB}0.069}            & \multicolumn{1}{c|}{$ 74 \pm 2$ ms} \\ \hline
        \end{tabular}
    }
    \caption{\textbf{Consistency analysis.} Results are color coded between \BGcolor{E0A4A4}{w}\BGcolor{E4B0A9}{o}\BGcolor{E9BCAE}{r}\BGcolor{EEC9B3}{s}\BGcolor{F3D5B8}{t}\BGcolor{F7E2BD}{ }\BGcolor{FCEEC2}{a}\BGcolor{F9F2C2}{n}\BGcolor{EBEBBA}{d}\BGcolor{DEE5B2}{ }\BGcolor{D1DEAA}{b}\BGcolor{C3D7A2}{e}\BGcolor{B6D09A}{s}\BGcolor{A8C992}{t}. 
    $^{\dag}$indicates training without normals and depth maps as input. The runtime is reported on an RTX 3090. For Kocsis et al., $^{\times N}$ indicates the number of predictions that are averaged, since this stochastic method is capable of sampling multiple predictions for a given input.
    Note how feeding diffusion hyperfeatures to the UNet architecture greatly improves multi-view consistency compared to RGB input, especially for roughness and metallic.
    }
    \label{tab:consistency}
\end{table}

\begin{figure*}[t]
    \centering
    \includegraphics[width=\textwidth]{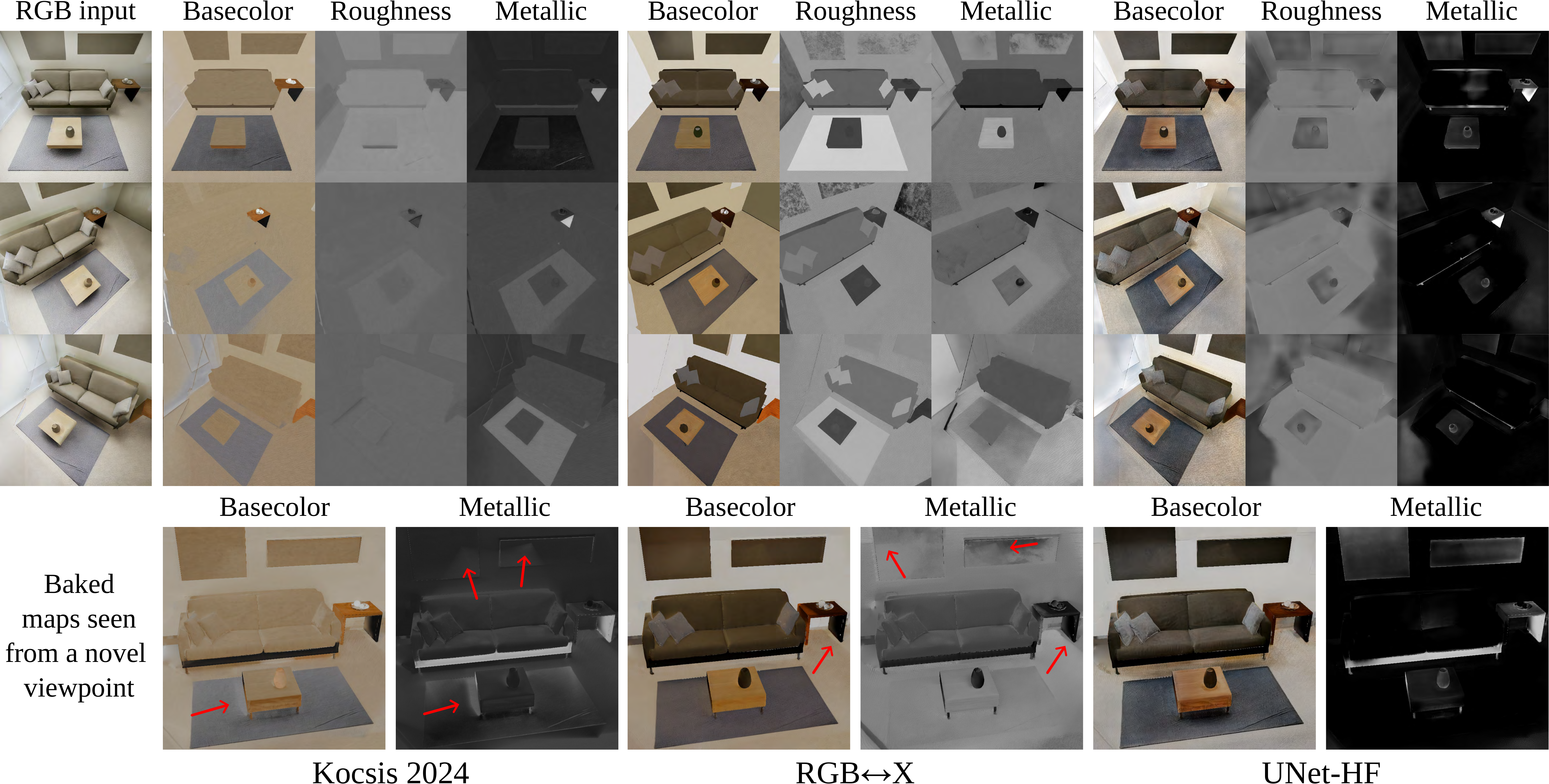}
    \caption{Qualitative comparison of single-image SVBRDF estimations merged in a texture atlas. On the left we show the input of three SVBRDF estimation techniques (Kocsis 2024, RGB2X, UNet-HF). The top row shows the SVBRDf predictions produced for each image. While Kocsis 2024 and RGB2X produce sharp, piecewise constant maps, these maps are not consistent across views, resulting in visible seams when merged in a texture atlas and rendered from a novel viewpoint (bottom row, red arrows). The UNet-HF architecture achieves higher consistency across views, which reduces the presence of seams (right).}
    \label{fig:consistency_baking}
\end{figure*}

\begin{figure*}[t]
    \centering
    \includegraphics[width=\textwidth]{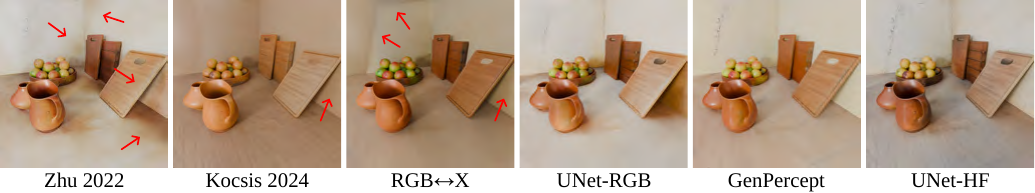}
    \caption{
    Rendered textured scenes using different SVBRDF estimators. \cite{zhu2022learning} produces sharp but splotchy maps, \cite{kocsis2024iid} generates monotonic maps lacking details, while \cite{RGB2X2024Zeng} generates blurry maps with visible seams due to the inconsistent estimations. UNet-RGB, UNet-HF and GenPercept produce more contrasted results with reduced seams artifacts, providing tradeoffs in terms of sharpness, colors and consistency. Please see the accompanying video for an animated result.
    }
    \label{fig:render_com}
\end{figure*}

\paragraph*{Multi-view consistency.}
Since our goal is to generate SVBRDF maps to texture a 3D scene, multi-view consistency is important to avoid artifacts in the merged texture atlas. Even if the SVBRDF maps of each image are plausible, if they are not consistent, seams and blur may appear in the texture-space material maps, which may negatively affect rendering quality. Fig.~\ref{fig:consistency_baking} and \ref{fig:render_com} point to such seams in the merged texture atlas obtained with \cite{kocsis2024iid} and \cite{RGB2X2024Zeng}.

\begin{figure}[t]
    \centering
    \includegraphics[width=0.5\textwidth]{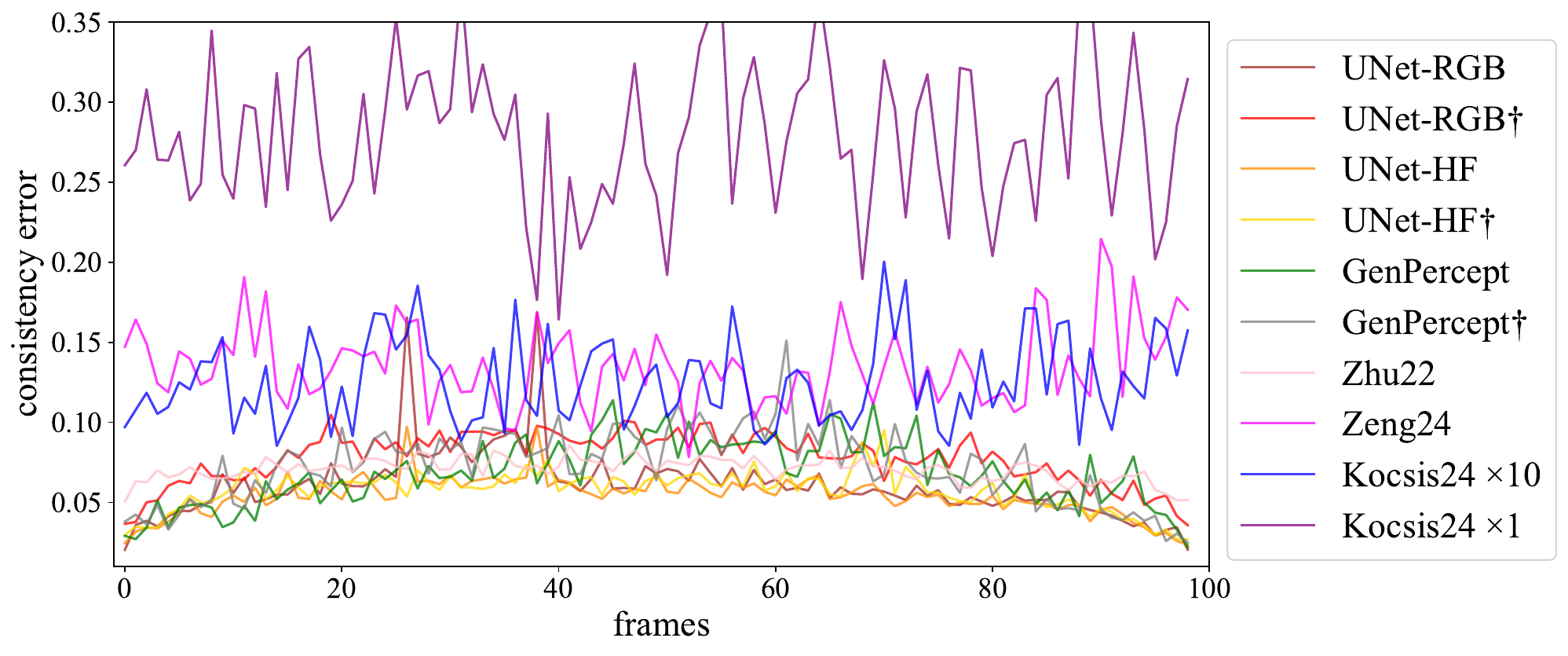}
    \caption{
    We show the Stop-the-Pop~\cite{radl2024stopthepop} metric along a video of the predicted maps for the synthetic scene 'kitchen-005', for the basecolor, on each method. UNet-HF is the most consistent across views (orange and yellow).
    }
    \label{fig:consistency_stop}
\end{figure}

For a qualitative evaluation of multi-view consistency, we find that short video segments of a fly through of the maps using different methods are particularly informative. We include such videos in our supplemental materials. These show that UNet-HF results in the lowest level of flickering. We evaluate this quantitatively using the flickering metric proposed by Stop-the-Pop~\cite{radl2024stopthepop} that warps each frame of the video to its subsequent frame and computes the \FLIP \cite{Andersson2020FLIP} difference between the two images.
As we have geometry and camera information, we use pixel-perfect depth maps and cameras for the warping, instead of an optical flow estimation as in Stop-the-Pop. We report in Tab.~\ref{tab:consistency} the sum of \FLIP metrics over 5 synthetic scenes that are not part of InteriorVerse, with 100 views each, warping only successive frames to evaluate short range flickering. We visualize the metric over time for a given scene in Fig.~\ref{fig:consistency_stop}.

This experiment reveals that the UNet-HF provides the best multi-view consistency overall. A possible interpretation of this result is that the deep features extracted from the image generation models have little dependence on viewpoint, yielding similar material values for objects seen in different views. Note also that averaging 10 predictions increases coherence for the method by Kocsis et al.~\cite{kocsis2024iid}, but in our experiments we noticed that this tends to reduce overall contrast and details.
The runtime column of Table~\ref{tab:consistency} illustrates that regression models (bottom section) are much faster than generative models (top section) because they only need a single inference through the architecture (compared to 50 inferences for diffusion models). The UNet-HF architecture accumulates activation features during RGB image generation, and hence requires only a single inference pass through the SVBRDF estimator. As for accuracy, providing additional geometric inputs to the predictors consistently improves the consistency of the estimations.

\paragraph*{Runtime analysis.} Table~\ref{tab:consistency} additionally provides the runtime of each predictor, generative methods on top and regression methods in the bottom. We see that the former \cite{kocsis2024iid, RGB2X2024Zeng} have runtimes of at least 6 seconds, making them unsuitable for a fast, interactive texturing pipeline. This is due to multiple factors: the sampling strategy which requires many iterative steps, the independent predictions of each channel for the method by Zeng et al.\cite{RGB2X2024Zeng}, and the averaging of multiple outputs for Kocsis et al.~\cite{kocsis2024iid}. On the other hand, regression-based methods (in the bottom part) only require a single inference making them more suitable for a fast texturing pipeline.

\section{Easy and Powerful 3D Scene Appearance Modeling}
\label{sec:results}
We now provide results of the multiview texturing pipeline described in Section~\ref{sec:pipeline}, which leverages image generation and SVBRDF estimation for appearance modeling of 3D scenes. We adopted the \textbf{UNet-HF} SVBRDF predictor to produce these results since this is the design that achieves the highest multi-view consistency with competitive single-view accuracy. 

\paragraph*{Implementation.}
We based our implementation of multi-view image generation on the Stable Diffusion ControlNet and ControlNet Inpainting pipelines from the Diffusers repository~\cite{vonplatenetal2022diffusers}. 
Specifically, we use Stable Diffusion v1.5 and the associated ControlNets (revision v1.1) \cite{controlnetv11} for contour, depth, and inpainting.
We generate images at $768^{2}$ resolution to obtain fine details in the textures. 

\begin{figure*}[!h]
    \centering
    \includegraphics[width=\textwidth]{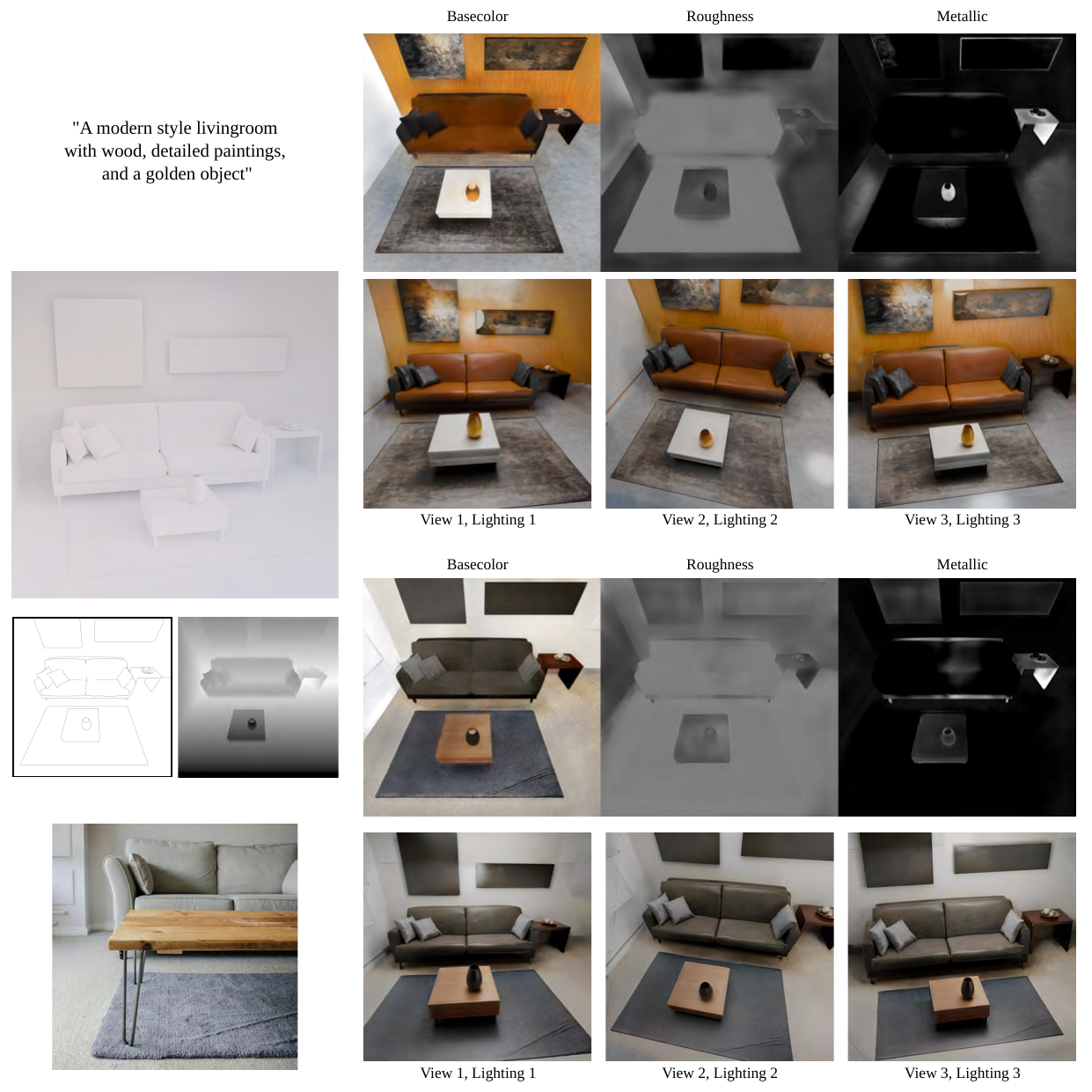}
    \caption{
    An example of material design for a living room scene. On the left, we show the two different prompts as well as images used to design the appearance of the scene (the middle image is a shaded version of the input geometry). For each prompt we show the material maps extracted from the generated image, and below the maps three different viewing and lighting conditions (moving light source).
    \label{fig:design_livingroom}
    }
\end{figure*}

\begin{figure*}[!h]
    \centering
    \includegraphics[width=\textwidth]{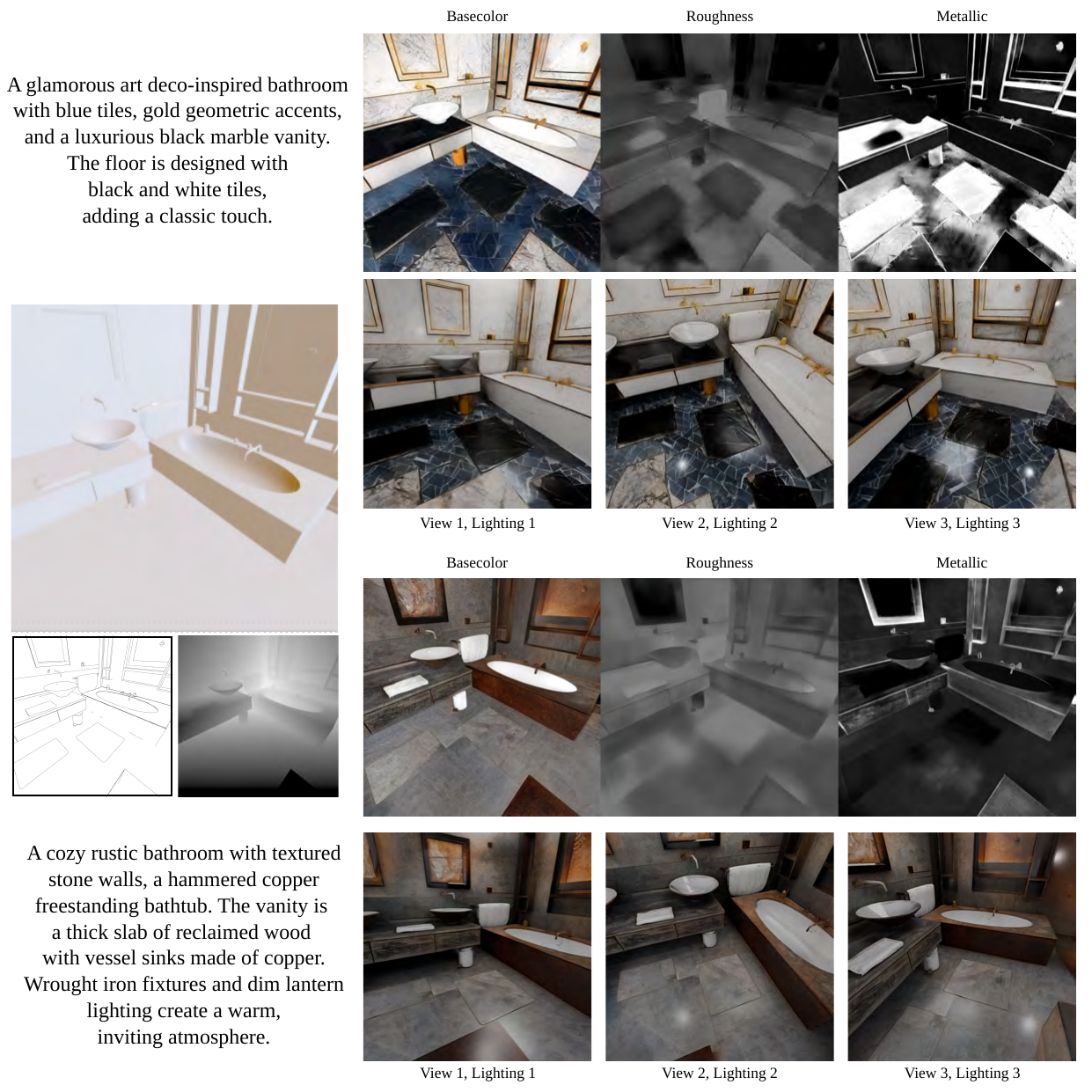}
    \caption{
    An example of material design for a bathroom scene. On the left, we show the two different prompts used to design the appearance of the scene. For each prompt we show the material maps extracted from the generated image, and below the maps three different viewing and lighting conditions (moving light source).
    \label{fig:design_bathroom}
    }
\end{figure*}

\begin{figure*}[!h]
    \centering
    \includegraphics[width=\textwidth]{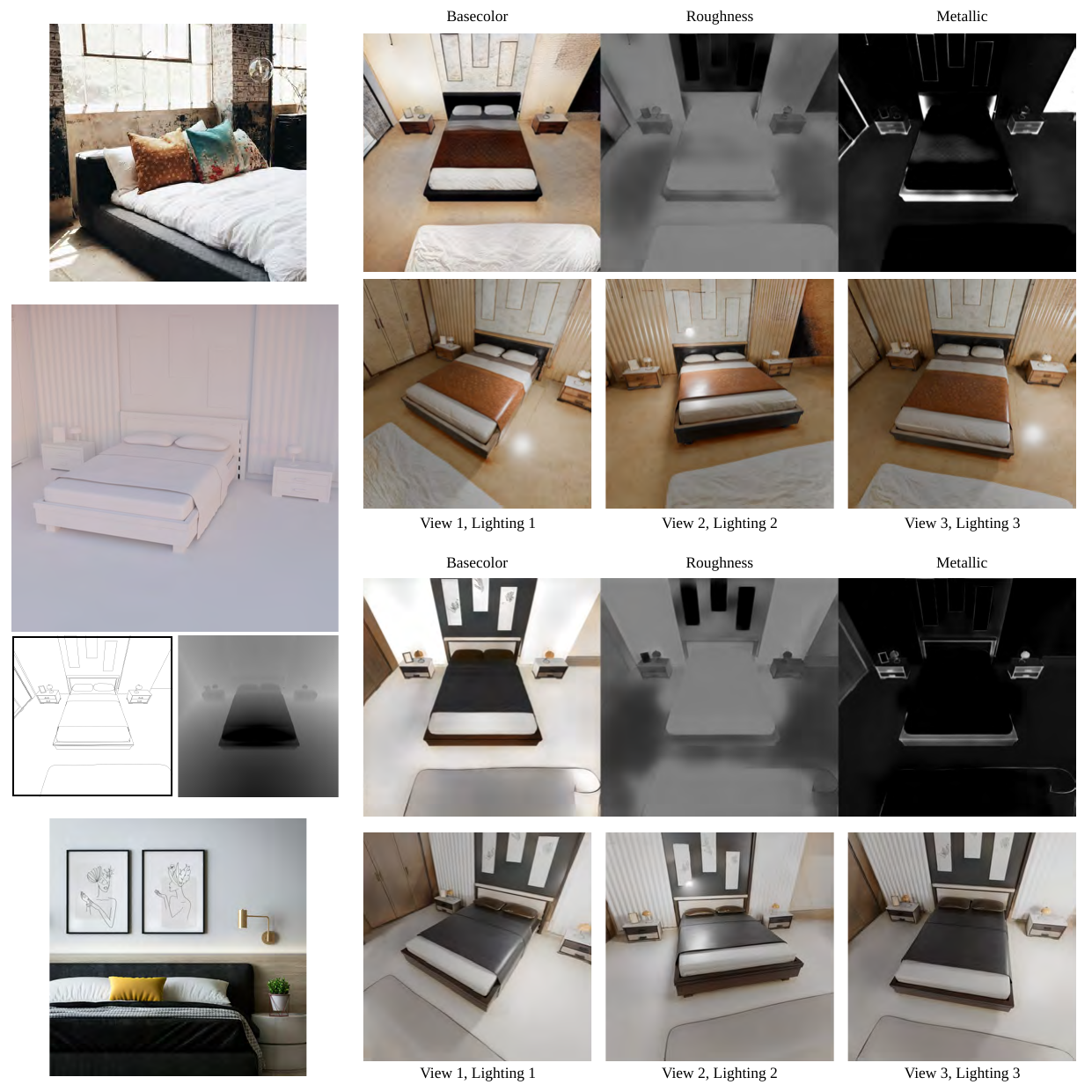}
    \caption{
    An example of material design for a bedroom scene. On the left, we show the two different image prompts used to design the appearance of the scene. For each prompt we show the material maps extracted from the generated image, and below the maps three different viewing and lighting conditions (moving light source).
    \label{fig:design_bedroom}
    }
\end{figure*}

\paragraph*{Results.}
Figures~\ref{fig:design_livingroom}~,~\ref{fig:design_bathroom}~and~\ref{fig:design_bedroom} show three scenes and different results conditioned either by text or image prompts. 
The generated SVBRDF atlases exhibit different materials, with a specular mineral floor and a rough carpet in Fig.~\ref{fig:design_livingroom}. A golden vase and a dark plastic vase are also displayed. 
Fig.~\ref{fig:design_bathroom} shows white, blue, and black marble, and thin golden structures, a stone wall, and copper furniture.
Finally, Fig.~\ref{fig:design_bedroom} shows a rough rug-like material and a specular white mineral for the floor. Additional texturing results are shown in the supplementary materials (file and video).

Being fast and image-based, this pipeline provides a convenient and easy way to design and iterate over the relightable appearance of an untextured 3D scene geometry. %
Since the first RGB image is generated in a few seconds, the user can easily iterate over the appearance of the first view. Similarly, if the user is not satisfied by the inpainting in a view, they can regenerate that view (though we did not use this feature in our results).
All relit results shown are computed from the final merged atlas, not from a single-view estimate. 
Note that the viewpoint used for relighting are different from the five views used for generation.
We include further results that illustrate the benefits of our appearance modeling pipeline in our supplemental video.%

\begin{figure}[t]
    \centering
    \includegraphics[width=0.47\textwidth]{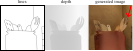}
    \caption{
    Thin objects may cause artifacts due to reprojection, as shown in the right part of the wooden fork, which does not generate clean material and geometry.
    \label{fig:thin_objects}
    }
\end{figure}

\section{Limitations and Future Work}
We base our study on the InteriorVerse dataset which, while being the only one available, has limitations in terms of diversity and precision of the ground truth PBR materials. 
It contains a small number of materials for floor, walls and furniture, and metallic objects are rare. In addition the distributions of roughness and metallic values are uneven (99\% of the roughness values are below 0.8). A richer training dataset should result in improved variety of generated materials.

We focused our study on single-image scene material estimation methods, which can still suffer from inconsistencies across viewpoints despite our careful choice of architecture and inputs. Multi-view SVBRDF extraction from photographs has been explored~\cite{DADDB19, choi2023mair}, but the generated images we are dealing with present illumination inconsistencies that might challenge such approaches. 

We demonstrated the complementarity of image generation and material prediction by implementing a fast pipeline for appearance modeling of 3D scenes. 
This pipeline suffers from a few limitations. First, the reprojection may cause artifacts in the presence of thin objects and small geometry, as illustrated in Fig.~\ref{fig:thin_objects}.
Second, as we rely on a diffusion model for appearance generation, and deep networks for material estimation, the expressivity and precision of our pipeline is limited by the quality of the models and training datasets. For example, we noticed that image inpainting sometimes produces visible seams, which might remain visible in the merged texture atlas.
Finally, using our prototype, a few iterations of prompt engineering can be required to achieve the desired appearance, which could be improved with the better prompt adherence of recent diffusion models~\cite{podell2023sdxl, Peebles2022DiT}.

\section{Conclusion}
In this paper we study the complementarity of generative image models and deep material predictors by evaluating the different choices required in a pipeline for designing materials over 3D scenes.
This pipeline takes the untextured geometry of a 3D scene as input and lets the designer provide additional conditions (e.g. text or image prompts). From this input, the pipeline quickly generates an SVBRDF texture atlas 
compatible with physically-based rendering. This simple pipeline enables rapid design iterations, and will directly benefit from the rapid progress in image generation control and quality.%

We study the impact of the type and input of neural architectures for SVBRDF estimation in the context of this 3D material texturing pipeline. Surprisingly, we find that a simple single-pass UNet architecture can outperform more complex recent solutions in terms of accuracy of prediction, even though recent methods can provide appealing piecewise constant results. 
We also note that %
the use of hyperfeatures improves multi-view consistency, possibly because they encode richer semantic information that is more invariant to view changes than RGB values.
Finally, we find that providing depth and normals as additional channels provides marginal results improvements.

\section{Acknowledgements}

We thank the anonymous reviewers for their valuable feedback. This work was funded by the European Research Council (ERC) Advanced Grant NERPHYS, number 101141721 https://project.inria.fr/nerphys. The authors are grateful to the OPAL infrastructure of the Université Côte d'Azur for providing resources and support, as well as Adobe and NVIDIA for software and hardware donations. This work was granted access to the HPC resources of IDRIS under the allocation AD011015561 made by GENCI. F. Durand acknowledges funding from from Google,  Amazon, and MIT-GIST.

\bibliographystyle{eg-alpha-doi} 
\bibliography{bibliography.bib}       

\newcommand{\etalchar}[1]{$^{#1}$}
\begin{thebibliography}{\uppercase{PWMAZ24}}

\bibitem[ALF22]{Avrahami_2022_CVPR}
\textsc{Avrahami O., Lischinski D., Fried O.}:
\newblock Blended diffusion for text-driven editing of natural images.
\newblock In \emph{Proc. IEEE Conference on Computer Vision and Pattern
  Recognition (CVPR)} (2022).

\bibitem[ALKN19]{azinovic2019inverse}
\textsc{Azinovic D., Li T.-M., Kaplanyan A., Nie{\ss}ner M.}:
\newblock Inverse path tracing for joint material and lighting estimation.
\newblock In \emph{Proceedings of the IEEE/CVF conference on computer vision
  and pattern recognition} (2019), pp.~2447--2456.

\bibitem[ANA{\etalchar{*}}20]{Andersson2020FLIP}
\textsc{Andersson P., Nilsson J., Akenine{-}M{\"{o}}ller T., Oskarsson M.,
  {\AA}str{\"{o}}m K., Fairchild M.~D.}:
\newblock {{\FLIP:} {A} Difference Evaluator for Alternating Images}.
\newblock \emph{Proceedings of the ACM on Computer Graphics and Interactive
  Techniques 3}, 2 (2020), 15:1--15:23.

\bibitem[BMHF23]{bhattad2023stylegan}
\textsc{Bhattad A., McKee D., Hoiem D., Forsyth D.}:
\newblock Stylegan knows normal, depth, albedo, and more.
\newblock In \emph{Advances in Neural Information Processing Systems} (2023).

\bibitem[BRV{\etalchar{*}}21]{baranchuk2021labelefficient}
\textsc{Baranchuk D., Rubachev I., Voynov A., Khrulkov V., Babenko A.}:
\newblock Label-efficient semantic segmentation with diffusion models, 2021.
\newblock \href {http://arxiv.org/abs/2112.03126} {\path{arXiv:2112.03126}}.

\bibitem[BS12]{burley2012physically}
\textsc{Burley B., Studios W. D.~A.}:
\newblock Physically-based shading at disney.
\newblock In \emph{ACM Siggraph} (2012), pp.~1--7.

\bibitem[CCCS08]{CCCS08}
\textsc{Callieri M., Cignoni P., Corsini M., Scopigno R.}:
\newblock Masked photo blending: mapping dense photographic dataset on
  high-resolution 3d models.
\newblock \emph{Computer \& Graphics 32}, 4 (Aug 2008), 464--473.

\bibitem[CDG{\etalchar{*}}24]{ceylan2024matatlas}
\textsc{Ceylan D., Deschaintre V., Groueix T., Martin R., Huang C.-H., Rouffet
  R., Kim V., Lassagne G.}:
\newblock Matatlas: Text-driven consistent geometry texturing and material
  assignment, 2024.

\bibitem[CLL{\etalchar{*}}24]{chen2024scenetex}
\textsc{Chen D.~Z., Li H., Lee H.-Y., Tulyakov S., Nie{\ss}ner M.}:
\newblock Scenetex: High-quality texture synthesis for indoor scenes via
  diffusion priors.
\newblock In \emph{Proceedings of the IEEE/CVF Conference on Computer Vision
  and Pattern Recognition} (2024), pp.~21081--21091.

\bibitem[CLP{\etalchar{*}}23]{choi2023mair}
\textsc{Choi J., Lee S., Park H., Jung S.-W., Kim I.-J., Cho J.}:
\newblock Mair: multi-view attention inverse rendering with 3d
  spatially-varying lighting estimation.
\newblock In \emph{2023 IEEE/CVF Conference on Computer Vision and Pattern
  Recognition (CVPR)} (2023), IEEE, pp.~8392--8401.

\bibitem[CSL{\etalchar{*}}23]{chen2023text2tex}
\textsc{Chen D.~Z., Siddiqui Y., Lee H.-Y., Tulyakov S., Nie{\ss}ner M.}:
\newblock Text2tex: Text-driven texture synthesis via diffusion models.
\newblock \emph{arXiv preprint arXiv:2303.11396} (2023).

\bibitem[CVW23]{chen2023beyond}
\textsc{Chen Y., Vi{\'e}gas F., Wattenberg M.}:
\newblock Beyond surface statistics: Scene representations in a latent
  diffusion model.
\newblock \emph{arXiv preprint arXiv:2306.05720} (2023).

\bibitem[DAD{\etalchar{*}}18]{DADDB18}
\textsc{Deschaintre V., Aittala M., Durand F., Drettakis G., Bousseau A.}:
\newblock Single-image svbrdf capture with a rendering-aware deep network.
\newblock \emph{ACM Trans. Graph. 37}, 4 (jul 2018).

\bibitem[DAD{\etalchar{*}}19]{DADDB19}
\textsc{Deschaintre V., Aittala M., Durand F., Drettakis G., Bousseau A.}:
\newblock Flexible svbrdf capture with a multi-image deep network.
\newblock \emph{Computer Graphics Forum(Eurographics Symposium on Rendering
  Conference Proceedings) 38}, 4 (jul 2019), 13.
\newblock URL: \url{http://www-sop.inria.fr/reves/Basilic/2019/DADDB19}.

\bibitem[DLG21]{deschaintre21}
\textsc{Deschaintre V., Lin Y., Ghosh A.}:
\newblock Deep polarization imaging for 3d shape and svbrdf acquisition.
\newblock In \emph{Proceedings of the IEEE/CVF Conference on Computer Vision
  and Pattern Recognition (CVPR)} (June 2021).

\bibitem[DOW{\etalchar{*}}24]{deng2024flashtex}
\textsc{Deng K., Omernick T., Weiss A., Ramanan D., Zhu J.-Y., Zhou T.,
  Agrawala M.}:
\newblock Flashtex: Fast relightable mesh texturing with lightcontrolnet.
\newblock In \emph{European Conference on Computer Vision (ECCV)} (2024).

\bibitem[FSW{\etalchar{*}}24]{fang2024makeitreal}
\textsc{Fang Y., Sun Z., Wu T., Wang J., Liu Z., Wetzstein G., Lin D.}:
\newblock Make-it-real: Unleashing large multimodal model for painting 3d
  objects with realistic materials, 2024.
\newblock \href {http://arxiv.org/abs/2404.16829} {\path{arXiv:2404.16829}}.

\bibitem[GSH{\etalchar{*}}20]{GuoMaterialGAN2020}
\textsc{Guo Y., Smith C., Ha\v{s}an M., Sunkavalli K., Zhao S.}:
\newblock Materialgan: Reflectance capture using a generative svbrdf model.
\newblock \emph{ACM Trans. Graph. 39}, 6 (nov 2020).

\bibitem[HCO{\etalchar{*}}23]{hoellein2023text2room}
\textsc{H\"ollein L., Cao A., Owens A., Johnson J., Nie{\ss}ner M.}:
\newblock Text2room: Extracting textured 3d meshes from 2d text-to-image
  models.
\newblock In \emph{Proceedings of the IEEE/CVF International Conference on
  Computer Vision (ICCV)} (October 2023), pp.~7909--7920.

\bibitem[HGY{\etalchar{*}}24]{hong2024supermat}
\textsc{Hong Y., Guo Y.-C., Yi R., Chen Y., Cao Y.-P., Ma L.}:
\newblock Supermat: Physically consistent pbr material estimation at
  interactive rates.
\newblock \emph{arXiv preprint arXiv:2411.17515} (2024).

\bibitem[HJA20]{ho2020denoising}
\textsc{Ho J., Jain A., Abbeel P.}:
\newblock Denoising diffusion probabilistic models.
\newblock \emph{Advances in neural information processing systems 33} (2020),
  6840--6851.

\bibitem[HSM{\etalchar{*}}24]{hedlin2023unsupervised}
\textsc{Hedlin E., Sharma G., Mahajan S., Isack H., Kar A., Tagliasacchi A., Yi
  K.~M.}:
\newblock Unsupervised semantic correspondence using stable diffusion.
\newblock In \emph{Proceedings of the 37th International Conference on Neural
  Information Processing Systems} (Red Hook, NY, USA, 2024), NIPS '23, Curran
  Associates Inc.

\bibitem[HWH{\etalchar{*}}24]{he2024neural}
\textsc{He Z., Wang T., Huang X., Pan X., Liu Z.}:
\newblock Neural lightrig: Unlocking accurate object normal and material
  estimation with multi-light diffusion.
\newblock \emph{arXiv preprint arXiv:2412.09593} (2024).

\bibitem[HWLW24]{huang2024materialanything}
\textsc{Huang X., Wang T., Liu Z., Wang Q.}:
\newblock Material anything: Generating materials for any 3d object via
  diffusion.
\newblock \emph{arXiv preprint arXiv:2411.15138} (2024).

\bibitem[HZG{\etalchar{*}}23]{hong2023lrm}
\textsc{Hong Y., Zhang K., Gu J., Bi S., Zhou Y., Liu D., Liu F., Sunkavalli
  K., Bui T., Tan H.}:
\newblock Lrm: Large reconstruction model for single image to 3d.
\newblock \emph{arXiv preprint arXiv:2311.04400} (2023).

\bibitem[JSR{\etalchar{*}}22]{Mitsuba3}
\textsc{Jakob W., Speierer S., Roussel N., Nimier-David M., Vicini D., Zeltner
  T., Nicolet B., Crespo M., Leroy V., Zhang Z.}:
\newblock Mitsuba 3 renderer, 2022.
\newblock https://mitsuba-renderer.org.

\bibitem[KSN24]{kocsis2024iid}
\textsc{Kocsis P., Sitzmann V., Niessner M.}:
\newblock Intrinsic image diffusion for indoor single-view material estimation.
\newblock \emph{Conference on Computer Vision and Pattern Recognition (CVPR)}
  (2024).

\bibitem[KSV{\etalchar{*}}23]{invs2023}
\textsc{Kant Y., Siarohin A., Vasilkovsky M., Guler R.~A., Ren J., Tulyakov S.,
  Gilitschenski I.}:
\newblock invs : Repurposing diffusion inpainters for novel view synthesis.
\newblock In \emph{SIGGRAPH Asia 2023 Conference Papers} (2023).

\bibitem[LADL18]{Li:2018:DMC}
\textsc{Li T.-M., Aittala M., Durand F., Lehtinen J.}:
\newblock Differentiable monte carlo ray tracing through edge sampling.
\newblock \emph{ACM Trans. Graph. (Proc. SIGGRAPH Asia) 37}, 6 (2018),
  222:1--222:11.

\bibitem[LDP{\etalchar{*}}23]{luo2023dhf}
\textsc{Luo G., Dunlap L., Park D.~H., Holynski A., Darrell T.}:
\newblock Diffusion hyperfeatures: Searching through time and space for
  semantic correspondence.
\newblock In \emph{Advances in Neural Information Processing Systems} (2023).

\bibitem[LDW{\etalchar{*}}24]{luo2024readoutguidance}
\textsc{Luo G., Darrell T., Wang O., Goldman D.~B., Holynski A.}:
\newblock Readout guidance: Learning control from diffusion features.

\bibitem[LGL{\etalchar{*}}25]{DiffusionRenderer}
\textsc{Liang R., Gojcic Z., Ling H., Munkberg J., Hasselgren J., Lin Z.-H.,
  Gao J., Keller A., Vijaykumar N., Fidler S., Wang Z.}:
\newblock Diffusionrenderer: Neural inverse and forward rendering with video
  diffusion models.
\newblock \emph{arXiv preprint arXiv: 2501.18590} (2025).

\bibitem[LGT{\etalchar{*}}23]{Lin_2023_CVPR}
\textsc{Lin C.-H., Gao J., Tang L., Takikawa T., Zeng X., Huang X., Kreis K.,
  Fidler S., Liu M.-Y., Lin T.-Y.}:
\newblock Magic3d: High-resolution text-to-3d content creation.
\newblock In \emph{Proc. IEEE/CVF Conference on Computer Vision and Pattern
  Recognition (CVPR)} (2023).

\bibitem[LSR{\etalchar{*}}20]{li2020inverse}
\textsc{Li Z., Shafiei M., Ramamoorthi R., Sunkavalli K., Chandraker M.}:
\newblock Inverse rendering for complex indoor scenes: Shape, spatially-varying
  lighting and svbrdf from a single image.
\newblock In \emph{Proc. IEEE Conference on Computer Vision and Pattern
  Recognition} (2020), pp.~2475--2484.

\bibitem[LWH{\etalchar{*}}23]{Liu2023Zero1to3ZO}
\textsc{Liu R., Wu R., Hoorick B.~V., Tokmakov P., Zakharov S., Vondrick C.}:
\newblock Zero-1-to-3: Zero-shot one image to 3d object.
\newblock \emph{2023 IEEE/CVF International Conference on Computer Vision
  (ICCV)} (2023), 9264--9275.

\bibitem[LXLW24]{liu2024text}
\textsc{Liu Y., Xie M., Liu H., Wong T.-T.}:
\newblock Text-guided texturing by synchronized multi-view diffusion.
\newblock In \emph{SIGGRAPH Asia 2024 Conference Papers} (2024), pp.~1--11.

\bibitem[LXR{\etalchar{*}}18]{li2018learning}
\textsc{Li Z., Xu Z., Ramamoorthi R., Sunkavalli K., Chandraker M.}:
\newblock Learning to reconstruct shape and spatially-varying reflectance from
  a single image.
\newblock \emph{ACM Transactions on Graphics (TOG) 37}, 6 (2018), 1--11.

\bibitem[MCP{\etalchar{*}}24]{Mitra2024}
\textsc{Mitra N.~J., Ceylan D., Patashnik O., CohenOr D., Guerrero P., Huang
  C.-H., Sung M.}:
\newblock Diffusion models for visual content creation.
\newblock In \emph{Siggraph Tutorial} (2024).

\bibitem[NDDJK21]{nimierdavid2021material}
\textsc{Nimier-David M., Dong Z., Jakob W., Kaplanyan A.}:
\newblock {Material and Lighting Reconstruction for Complex Indoor Scenes with
  Texture-space Differentiable Rendering}.
\newblock In \emph{Eurographics Symposium on Rendering - DL-only Track} (2021),
  The Eurographics Association.

\bibitem[PEL{\etalchar{*}}23]{podell2023sdxl}
\textsc{Podell D., English Z., Lacey K., Blattmann A., Dockhorn T., M{\"u}ller
  J., Penna J., Rombach R.}:
\newblock Sdxl: Improving latent diffusion models for high-resolution image
  synthesis.
\newblock \emph{arXiv preprint arXiv:2307.01952} (2023).

\bibitem[PJBM23]{poole2023dreamfusion}
\textsc{Poole B., Jain A., Barron J.~T., Mildenhall B.}:
\newblock Dreamfusion: Text-to-3d using 2d diffusion.
\newblock In \emph{The Eleventh International Conference on Learning
  Representations} (2023).

\bibitem[PTL{\etalchar{*}}23]{pan2023_DragGAN}
\textsc{Pan X., Tewari A., Leimk\"{u}hler T., Liu L., Meka A., Theobalt C.}:
\newblock Drag your gan: Interactive point-based manipulation on the generative
  image manifold.
\newblock In \emph{ACM SIGGRAPH 2023 Conference Proceedings} (New York, NY,
  USA, 2023), SIGGRAPH '23, Association for Computing Machinery.

\bibitem[PWMAZ24]{perla2024easitex}
\textsc{Perla S. R.~K., Wang Y., Mahdavi-Amiri A., Zhang H.}:
\newblock Easi-tex: Edge-aware mesh texturing from single image.
\newblock \emph{ACM Transactions on Graphics (Proceedings of SIGGRAPH) 43}, 4
  (2024).

\bibitem[PX22]{Peebles2022DiT}
\textsc{Peebles W., Xie S.}:
\newblock Scalable diffusion models with transformers.
\newblock \emph{arXiv preprint arXiv:2212.09748} (2022).

\bibitem[PYG{\etalchar{*}}24]{STARDiffusion}
\textsc{Po R., Yifan W., Golyanik V., Aberman K., Barron J.~T., Bermano A.,
  Chan E., Dekel T., Holynski A., Kanazawa A., Liu C., Liu L., Mildenhall B.,
  Nießner M., Ommer B., Theobalt C., Wonka P., Wetzstein G.}:
\newblock State of the art on diffusion models for visual computing.
\newblock \emph{Computer Graphics Forum 43}, 2 (2024).

\bibitem[RBL{\etalchar{*}}22]{Rombach_2022_CVPR}
\textsc{Rombach R., Blattmann A., Lorenz D., Esser P., Ommer B.}:
\newblock High-resolution image synthesis with latent diffusion models.
\newblock In \emph{Proceedings of the IEEE/CVF Conference on Computer Vision
  and Pattern Recognition (CVPR)} (June 2022).

\bibitem[RDN{\etalchar{*}}22]{ramesh2022hierarchical}
\textsc{Ramesh A., Dhariwal P., Nichol A., Chu C., Chen M.}:
\newblock Hierarchical text-conditional image generation with clip latents.
\newblock \emph{arXiv preprint arXiv:2204.06125 1}, 2 (2022), 3.

\bibitem[RH01]{Ramamoorthi2001}
\textsc{Ramamoorthi R., Hanrahan P.}:
\newblock A signal-processing framework for inverse rendering.
\newblock In \emph{SIGGRAPH} (2001).

\bibitem[RMA{\etalchar{*}}23]{Richardson2023TEXTure}
\textsc{Richardson E., Metzer G., Alaluf Y., Giryes R., Cohen-Or D.}:
\newblock Texture: Text-guided texturing of 3d shapes.
\newblock In \emph{ACM SIGGRAPH 2023 Conference Proceedings} (New York, NY,
  USA, 2023), SIGGRAPH '23, Association for Computing Machinery.

\bibitem[RSP{\etalchar{*}}24]{radl2024stopthepop}
\textsc{Radl L., Steiner M., Parger M., Weinrauch A., Kerbl B., Steinberger
  M.}:
\newblock {StopThePop: Sorted Gaussian Splatting for View-Consistent Real-time
  Rendering}.
\newblock \emph{ACM Transactions on Graphics 43}, 4 (2024).

\bibitem[SME21]{song2021denoising}
\textsc{Song J., Meng C., Ermon S.}:
\newblock Denoising diffusion implicit models.
\newblock In \emph{International Conference on Learning Representations}
  (2021).

\bibitem[TZC{\etalchar{*}}23]{Tang2023mvdiffusion}
\textsc{Tang S., Zhang F., Chen J., Wang P., Furukawa Y.}:
\newblock Mvdiffusion: Enabling holistic multi-view image generation with
  correspondence-aware diffusion.
\newblock \emph{arXiv} (2023).

\bibitem[VBP{\etalchar{*}}24]{vainer2024collaborative}
\textsc{Vainer S., Boss M., Parger M., Kutsy K., De~Nigris D., Rowles C.,
  Perony N., Donn{\'e} S.}:
\newblock Collaborative control for geometry-conditioned pbr image generation.
\newblock \emph{arXiv preprint arXiv:2402.05919} (2024).

\bibitem[VKDN{\etalchar{*}}24]{vainer2024jointly}
\textsc{Vainer S., Kutsy K., De~Nigris D., Rowles C., Elizarov S., Donn{\'e}
  S.}:
\newblock Jointly generating multi-view consistent pbr textures using
  collaborative control.
\newblock \emph{arXiv preprint arXiv:2410.06985} (2024).

\bibitem[VMR{\etalchar{*}}23]{vecchio2023controlmat}
\textsc{Vecchio G., Martin R., Roullier A., Kaiser A., Rouffet R., Deschaintre
  V., Boubekeur T.}:
\newblock Controlmat: Controlled generative approach to material capture.
\newblock \emph{arXiv preprint arXiv:2309.01700} (2023).

\bibitem[vPPL{\etalchar{*}}22]{vonplatenetal2022diffusers}
\textsc{von Platen P., Patil S., Lozhkov A., Cuenca P., Lambert N., Rasul K.,
  Davaadorj M., Nair D., Paul S., Berman W., Xu Y., Liu S., Wolf T.}:
\newblock Diffusers: State-of-the-art diffusion models.
\newblock \url{https://github.com/huggingface/diffusers}, 2022.

\bibitem[WTC{\etalchar{*}}25]{wang2025materialist}
\textsc{Wang L., Tran D.~M., Cui R., TG T., Chandraker M., Frisvad J.~R.}:
\newblock Materialist: Physically based editing using single-image inverse
  rendering.
\newblock \emph{arXiv preprint arXiv:2501.03717} (2025).

\bibitem[WXM{\etalchar{*}}24]{wang2024boosting}
\textsc{Wang Y., Xu X., Ma L., Wang H., Dai B.}:
\newblock Boosting 3d object generation through pbr materials.
\newblock In \emph{SIGGRAPH Asia 2024 Conference Papers} (2024), pp.~1--11.

\bibitem[WZB{\etalchar{*}}24]{wei2024meshlrm}
\textsc{Wei X., Zhang K., Bi S., Tan H., Luan F., Deschaintre V., Sunkavalli
  K., Su H., Xu Z.}:
\newblock Meshlrm: Large reconstruction model for high-quality mesh.
\newblock \emph{arXiv preprint arXiv:2404.12385} (2024).

\bibitem[WZY{\etalchar{*}}23]{fipt2023}
\textsc{Wu L., Zhu R., Yaldiz M.~B., Zhu Y., Cai H., Matai J., Porikli F., Li
  T.-M., Chandraker M., Ramamoorthi R.}:
\newblock Factorized inverse path tracing for efficient and accurate
  material-lighting estimation.
\newblock In \emph{Proceedings of the IEEE/CVF International Conference on
  Computer Vision} (2023), pp.~3848--3858.

\bibitem[XGL{\etalchar{*}}25]{xu2024genpercept}
\textsc{Xu G., Ge Y., Liu M., Fan C., Xie K., Zhao Z., Chen H., Shen C.}:
\newblock What matters when repurposing diffusion models for general dense
  perception tasks?
\newblock In \emph{Proc. of the IEEE International Conf. on Learning
  Representations} (2025).

\bibitem[YHK{\etalchar{*}}24]{TextureDreamer2024CVPR}
\textsc{Yeh Y.-Y., Huang J.-B., Kim C., Xiao L., Nguyen-Phuoc T., Khan N.,
  Zhang C., Chandraker M., Marshall C.~S., Dong Z., Li Z.}:
\newblock Texturedreamer: Image-guided texture synthesis through geometry-aware
  diffusion.
\newblock In \emph{Proceedings of the IEEE/CVF Conference on Computer Vision
  and Pattern Recognition (CVPR)} (June 2024), pp.~4304--4314.

\bibitem[YLH{\etalchar{*}}23]{Yan:2023:PSDR-Room}
\textsc{Yan K., Luan F., Ha\v{s}an M., Groueix T., Deschaintre V., Zhao S.}:
\newblock Psdr-room: Single photo to scene using differentiable rendering.
\newblock In \emph{ACM SIGGRAPH Asia 2023 Conference Proceedings} (2023).

\bibitem[YOPM24]{youwang2024paintit}
\textsc{Youwang K., Oh T.-H., Pons-Moll G.}:
\newblock Paint-it: Text-to-texture synthesis via deep convolutional texture
  map optimization and physically-based rendering.
\newblock In \emph{IEEE Conference on Computer Vision and Pattern Recognition
  (CVPR)} (2024).

\bibitem[YZL{\etalchar{*}}23]{ye2023ipadapter}
\textsc{Ye H., Zhang J., Liu S., Han X., Yang W.}:
\newblock Ip-adapter: Text compatible image prompt adapter for text-to-image
  diffusion models.
\newblock 2023.

\bibitem[ZCQ{\etalchar{*}}24]{zeng2024paint3d}
\textsc{Zeng X., Chen X., Qi Z., Liu W., Zhao Z., Wang Z., Fu B., Liu Y., Yu
  G.}:
\newblock Paint3d: Paint anything 3d with lighting-less texture diffusion
  models.
\newblock In \emph{Proceedings of the IEEE/CVF conference on computer vision
  and pattern recognition} (2024), pp.~4252--4262.

\bibitem[ZDG{\etalchar{*}}24]{RGB2X2024Zeng}
\textsc{Zeng Z., Deschaintre V., Georgiev I., Hold-Geoffroy Y., Hu Y., Luan F.,
  Yan L.-Q., Ha\v{s}an M.}:
\newblock Rgb↔x: Image decomposition and synthesis using material- and
  lighting-aware diffusion models.
\newblock In \emph{ACM SIGGRAPH 2024 Conference Papers} (New York, NY, USA,
  2024), SIGGRAPH '24, Association for Computing Machinery.

\bibitem[ZIE{\etalchar{*}}18]{zhang2018perceptual}
\textsc{Zhang R., Isola P., Efros A.~A., Shechtman E., Wang O.}:
\newblock The unreasonable effectiveness of deep features as a perceptual
  metric.
\newblock In \emph{CVPR} (2018).

\bibitem[ZLG{\etalchar{*}}21]{zhang2021datasetgan}
\textsc{Zhang Y., Ling H., Gao J., Yin K., Lafleche J.-F., Barriuso A.,
  Torralba A., Fidler S.}:
\newblock Datasetgan: Efficient labeled data factory with minimal human effort.
\newblock In \emph{Proceedings of the IEEE/CVF Conference on Computer Vision
  and Pattern Recognition} (2021), pp.~10145--10155.

\bibitem[ZLH{\etalchar{*}}22]{zhu2022learning}
\textsc{Zhu J., Luan F., Huo Y., Lin Z., Zhong Z., Xi D., Wang R., Bao H.,
  Zheng J., Tang R.}:
\newblock Learning-based inverse rendering of complex indoor scenes with
  differentiable monte carlo raytracing.
\newblock In \emph{SIGGRAPH Asia 2022 Conference Papers} (2022), ACM.

\bibitem[ZLX{\etalchar{*}}24]{DreamMat2024Zhang}
\textsc{Zhang Y., Liu Y., Xie Z., Yang L., Liu Z., Yang M., Zhang R., Kou Q.,
  Lin C., Wang W., Jin X.}:
\newblock Dreammat: High-quality pbr material generation with geometry- and
  light-aware diffusion models.
\newblock \emph{ACM Transactions on Graphics (Proc. SIGGRAPH) 43}, 4 (jul
  2024).

\bibitem[ZPX{\etalchar{*}}24]{zhang2024mapa}
\textsc{Zhang S., Peng S., Xu T., Yang Y., Chen T., Xue N., Shen Y., Bao H., Hu
  R., Zhou X.}:
\newblock Mapa: Text-driven photorealistic material painting for 3d shapes.
\newblock In \emph{ACM SIGGRAPH 2024 Conference Papers} (2024), pp.~1--12.

\bibitem[ZPZ{\etalchar{*}}24]{TexPainter2024Zhang}
\textsc{Zhang H., Pan Z., Zhang C., Zhu L., Gao X.}:
\newblock Texpainter: Generative mesh texturing with multi-view consistency.
\newblock In \emph{ACM SIGGRAPH 2024 Conference Papers} (New York, NY, USA,
  2024), SIGGRAPH '24, Association for Computing Machinery.

\bibitem[ZRA23a]{zhang2023adding}
\textsc{Zhang L., Rao A., Agrawala M.}:
\newblock Adding conditional control to text-to-image diffusion models, 2023.

\bibitem[ZRA23b]{controlnetv11}
\textsc{Zhang L., Rao A., Agrawala M.}:
\newblock Controlnet-v1-1.
\newblock \url{https://github.com/lllyasviel/ControlNet-v1-1-nightly}, 2023.

\bibitem[ZRL{\etalchar{*}}23]{zhao2023unleashing}
\textsc{Zhao W., Rao Y., Liu Z., Liu B., Zhou J., Lu J.}:
\newblock Unleashing text-to-image diffusion models for visual perception.
\newblock \emph{ICCV} (2023).

\bibitem[ZWZ{\etalchar{*}}24]{zhang2024clay}
\textsc{Zhang L., Wang Z., Zhang Q., Qiu Q., Pang A., Jiang H., Yang W., Xu L.,
  Yu J.}:
\newblock Clay: A controllable large-scale generative model for creating
  high-quality 3d assets.
\newblock \emph{ACM Transactions on Graphics (TOG) 43}, 4 (2024), 1--20.

\bibitem[ZXS{\etalchar{*}}25]{zhang2025instex}
\textsc{Zhang Y., Xiong Z., Shen Z., Lin G., Wang H., Vun N.}:
\newblock Instex: Indoor scenes stylized texture synthesis.
\newblock \emph{arXiv preprint arXiv:2501.13969} (2025).

\end{thebibliography}

\appendix

\begin{figure*}[ht]
 \includegraphics[width=\textwidth]{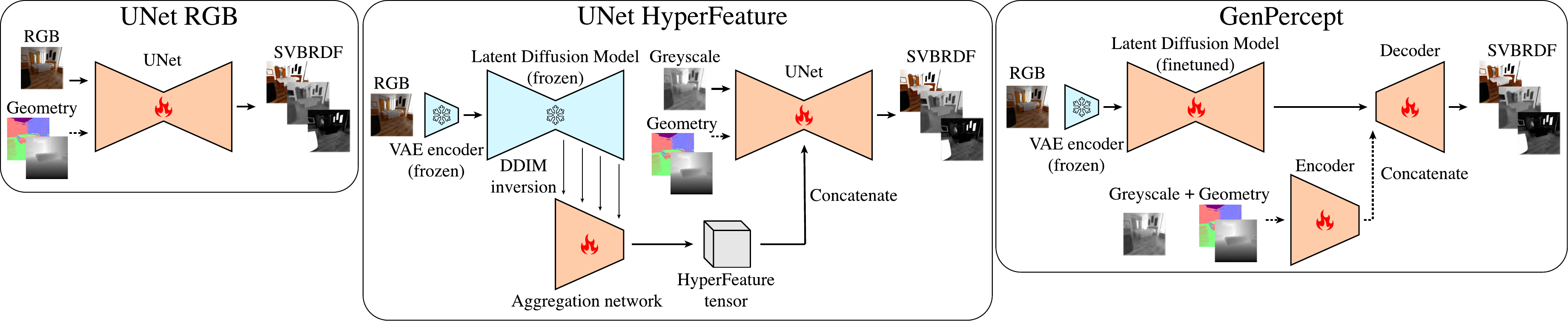}
 \centering
  \caption{
    Architectures used in our evaluation. The dotted line box indicates the optional inputs to each architecture (geometry and grayscale input).
    }
\label{fig:multiarchi}
\end{figure*}

\section{SVBRDF Predictors architectures}

\subsection{UNet-RGB}

For the UNet-RGB predictor (Fig.\ref{fig:multiarchi}, left), we base our implementation on the diffusers library \cite{vonplatenetal2022diffusers}. This codebase provides us with the implementation of the denoising U-Net of Stable Diffusion. For our predictor, we replace the default "AttnDownBlock2D" and "AttnUpBlock2D" by simpler "DownBlock2D" and "UpBlock2D" for a more efficient training. We use a Group Norm value of 1 ("norm\_num\_groups") as it helps to stabilize training. We leave the rest of the parameters to their default values. When trained solely on single images, the network (UNet-RGB$^{\dag}$) is provided with the tonemapped, sRGB corrected image (renormalized in [-1,1]) as input and directly outputs a 5-channel image containing basecolor, roughness, and metallic. When trained with additional geometry cues (depth and normals), the UNet-RGB takes as input a 7-channel image formed by the concatenation of the RGB image, the normalized depth and the normal map.

\subsection{UNet-HF}

For the UNet-HF predictor (Fig.\ref{fig:multiarchi}, middle), we extract activation features of Stable Diffusion's denoising U-Net during inverse DDIM sampling \cite{song2021denoising}, similar to \cite{luo2023dhf}. The author leverages an aggregation network, which processes upsampled activation features to the size of the biggest feature map in the decoder layers. This results in what they call a 'Hyperfeature tensor'.
The UNet-HF predictor shares the same U-Net architecture as the UNet-RGB network but mixes its middle bottleneck layer with the so-called 'Hyperfeature tensor'. To do so, we encode the Hyperfeature tensor using a Conv2D block with TanH activation, inspired by \cite{luo2024readoutguidance}, at half the original middle bottleneck depth resolution, and encode the original middle bottleneck of the U-Net with a similar Conv2D block with TanH activation. Both outputs are concatenated to produce a middle block of the same size as the original middle bottleneck. 
The aggregation network of \cite{luo2023dhf} and the prediction UNet are trained end-to-end, to extract the relevant features for SVBRDF prediction. We extract the diffusion features of a given RGB image by inverting it into Stable Diffusion using inverse DDIM sampling. Contrary to the UNet-RGB, we only provide the SVBRDF predictor with a grayscale version of the input image, forcing it to extract colored information from the Hyperfeature tensor rather than the image itself. We train two versions: UNet-HF, trained with input depth and normals, and UNet-HF$^{\dag}$, trained only on grayscale images.

Because we use convolutional neural networks, both UNet-RGB, and UNet-HF can predict SVBRDF at any resolution, when provided with the appropriate maps as input (the same resolution as the input RGB). Since the size of the Hyperfeature tensor is fixed to a height and width of size 64*64, we simply interpolate it to the same size as the bottleneck layer of the SVBRDF predictor to output maps at the appropriate resolution.

\subsection{GenPercept}

For the GenPercept predictor (Fig.\ref{fig:multiarchi}, right), we use the implementation of the original paper \cite{xu2024genpercept} using the customized decoder variant to account for our 5-channels outputs. For the decoder architecture, we simply output 5 channels out of the DPT head for SVBRDF estimation instead of a single one for monocular depth estimation. When trained with additional geometry cues (depth and normals), we first use a fully connected geometry head with ReLU activations to go from 5, to 16, to 32 features. We use a second head for the output of the diffusion UNet, taken from the original \cite{xu2024genpercept} codebase, to which we remove the last Conv2D layer. Finally, we concatenate both 32 features of the geometry head and the former one, to pass them into a two-layered fully connected output head (64, to 32, to 5 features) with a ReLU activation in between and an Identity activation in the end (to stay as close as possible to the original implementation).
For training, we use an 'effective\_batch\_size' of 32 and a 'max\_train\_batch\_size' of 8. Both UNet and decoder learning rates are set to 3e-5. We set 'max\_epoch' to 10.000 and the 'lr\_scheduler' to 'IterExponential' with parameters total\_iter=25000, final\_ratio=0.01, and warmup\_steps=100.

\section{Additional Texturing result}

\begin{figure*}[ht]
      \centering
      \includegraphics[width=\textwidth]{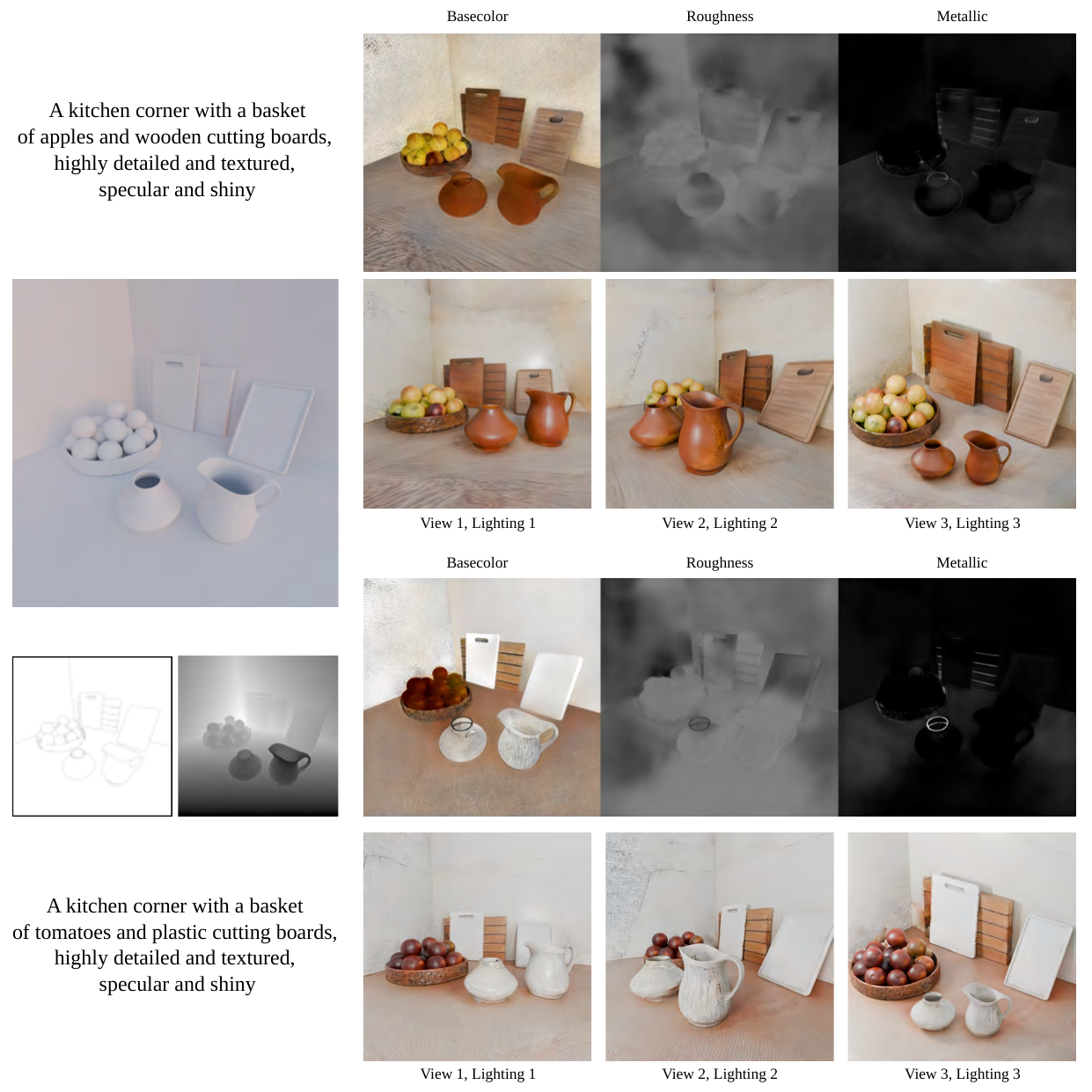}
      \caption{An example of material design for a kitchen scene. On the left, we show the two different prompts used to design the appearance of the scene. For each prompt we show the material maps extracted from the generated image, and below the maps three different viewing and lighting conditions (moving lightsource).}
      \label{fig:design_kitchen}
\end{figure*}

In Fig.\ref{fig:design_kitchen}, we illustrate another texturing example on a kitchen scene with a basket, some fruits, cutting boards and pots.

\section{Additional Stop-The-Pop metrics}

\begin{figure*}
    \centering
    \begin{subfigure}{0.25\textwidth}
        \centering
        \includegraphics[width=\linewidth]{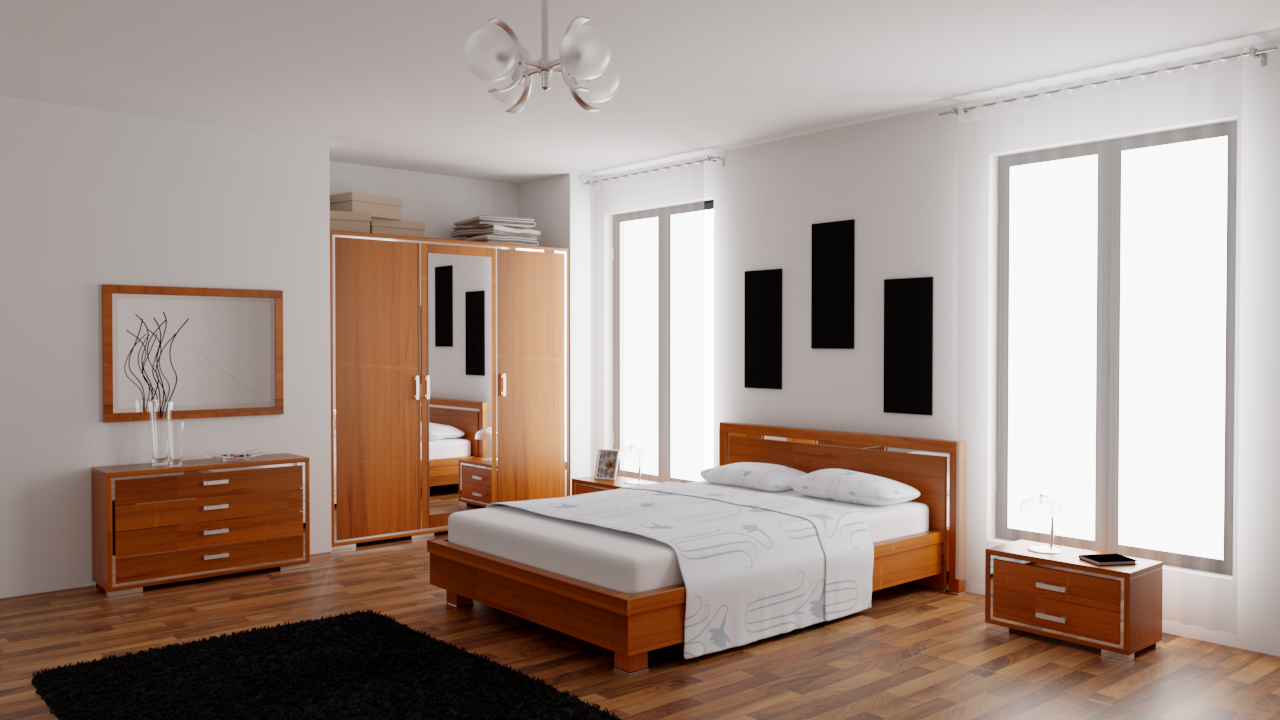}
        \caption{bedroom scene}
        \label{fig:sfig1}
    \end{subfigure}
    \hfil
    \begin{subfigure}{0.2\textwidth}
        \centering
        \includegraphics[width=\linewidth]{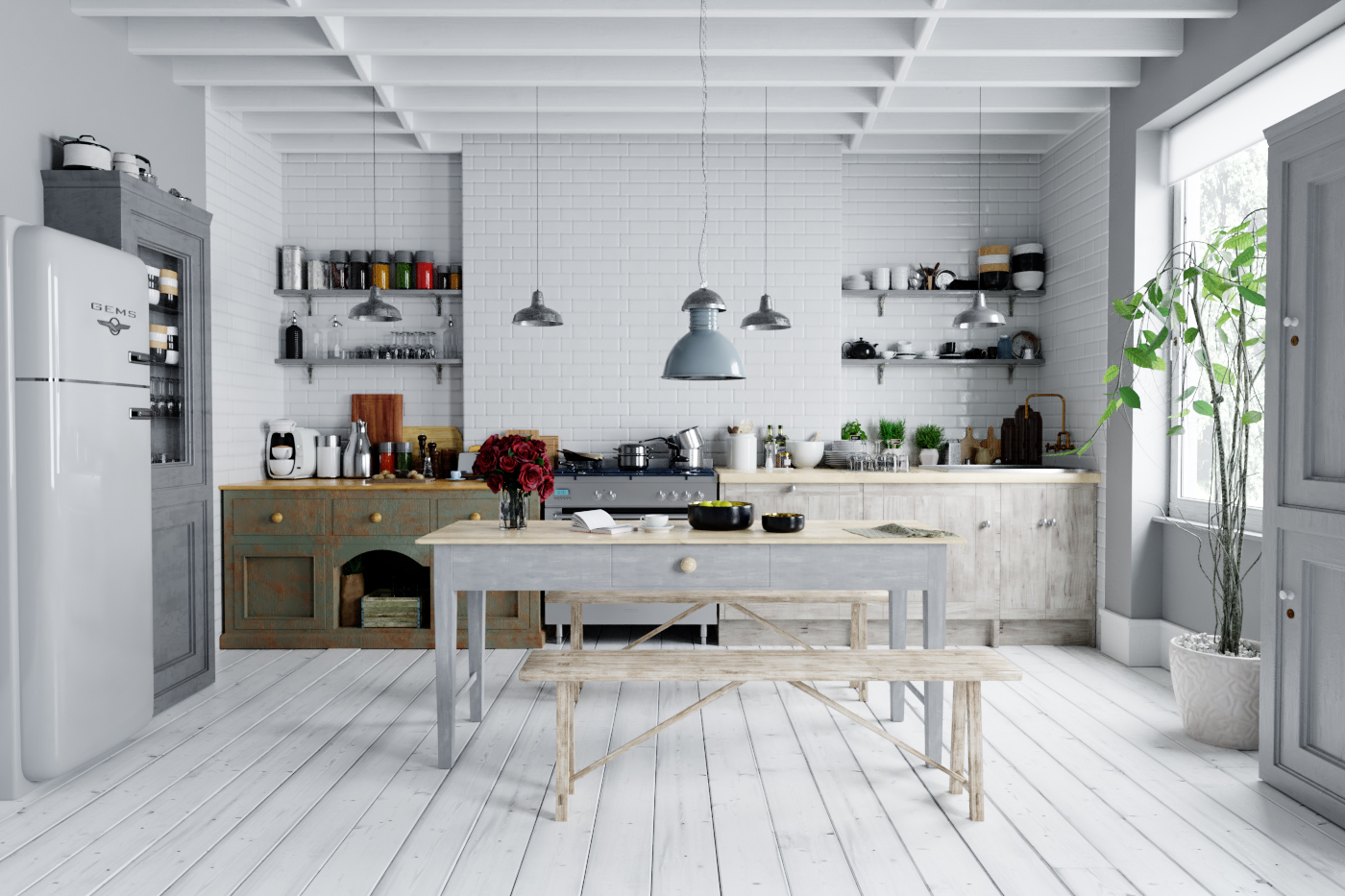}
        \caption{'kitchen-003' scene}
        \label{fig:sfig2}
    \end{subfigure}
    \hfil
    \begin{subfigure}{0.2\textwidth}
        \centering
        \includegraphics[width=\linewidth]{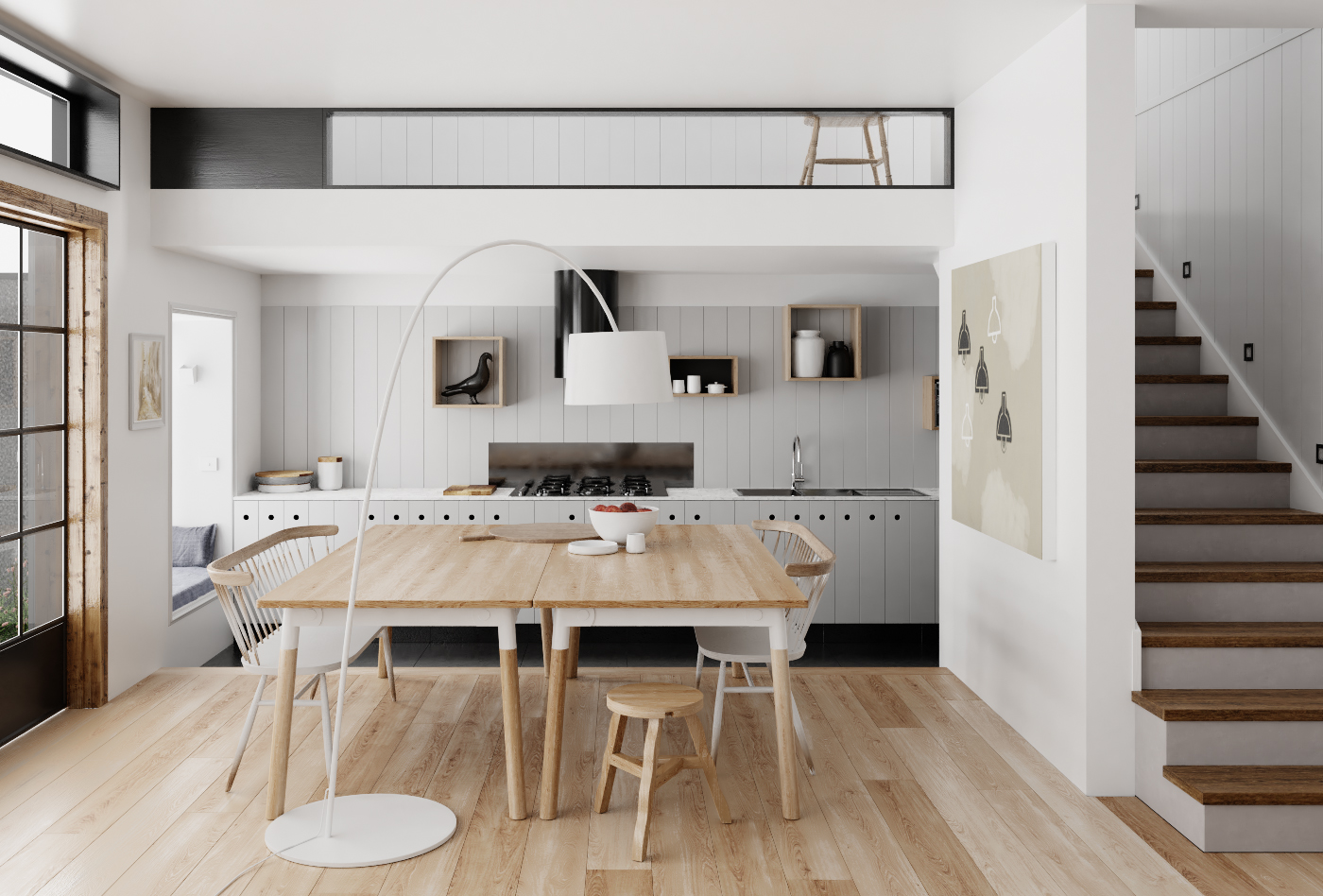}
        \caption{'kitchen-005' scene}
        \label{fig:sfig2} 
    \end{subfigure}
    
    \begin{subfigure}{.2\textwidth}
        \centering
        \includegraphics[width=\linewidth]{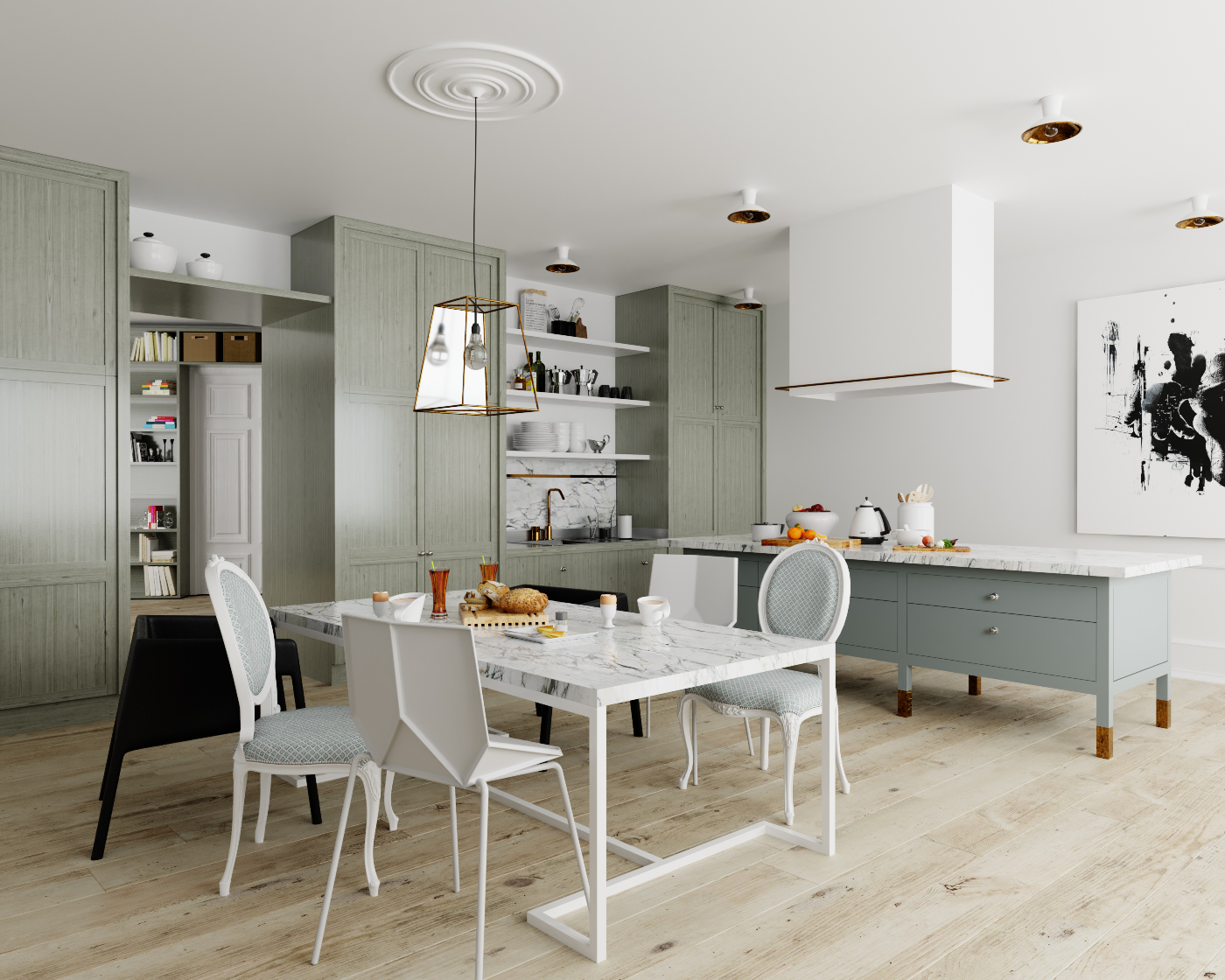}
        \caption{'kitchen-007' scene}
        \label{fig:sfig2}
    \end{subfigure}
    \hfil
    \begin{subfigure}{.2\textwidth}
        \centering
        \includegraphics[width=\linewidth]{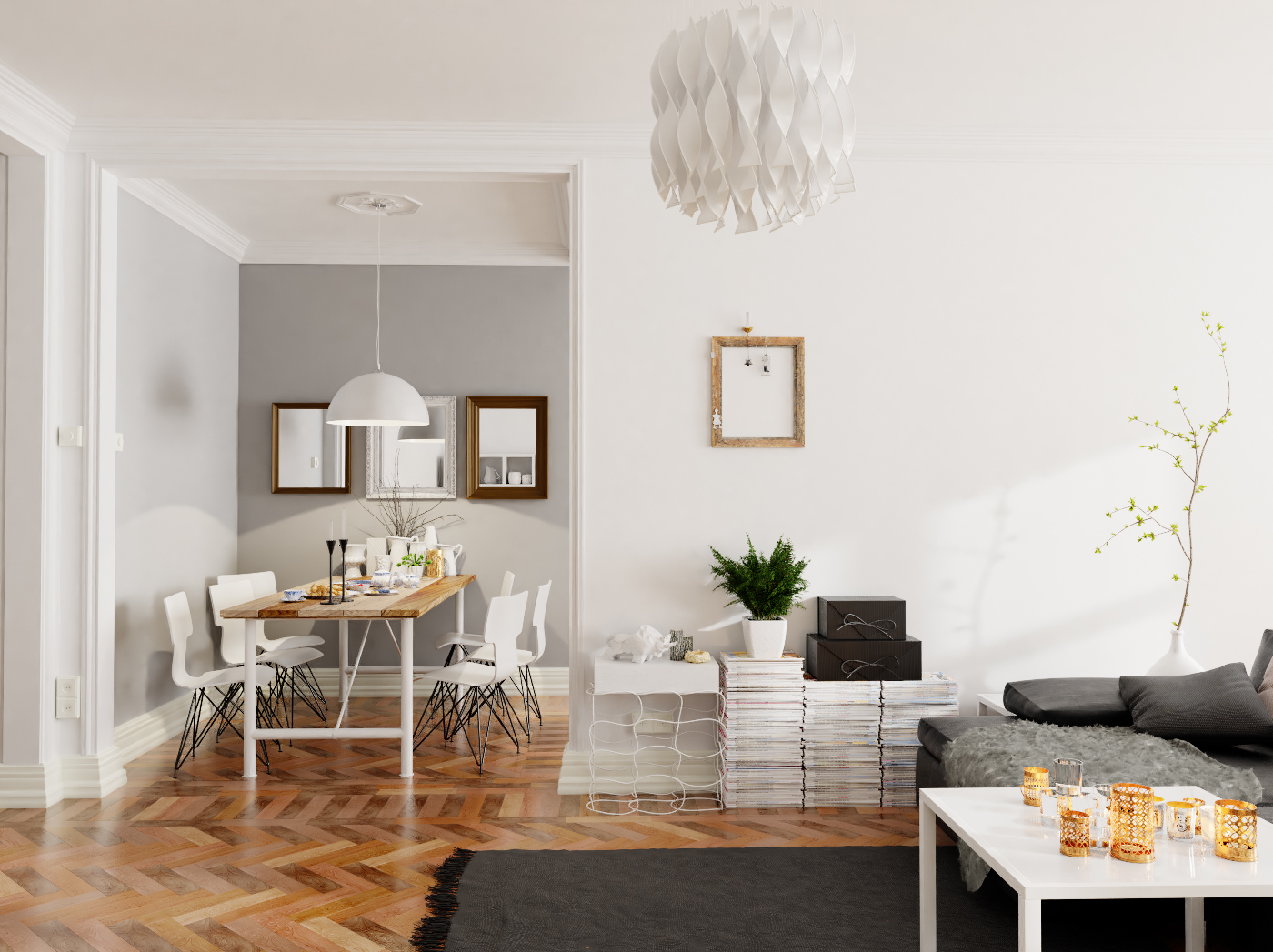}
        \caption{'kitchen-010' scene}
        \label{fig:sfig2}
    \end{subfigure}
    \caption{Scenes used to compute the Stop-The-Pop metrics.}
    \label{fig:five_scenes}
\end{figure*}

\begin{figure*}[ht]
      \centering
      \includegraphics[width=\textwidth]{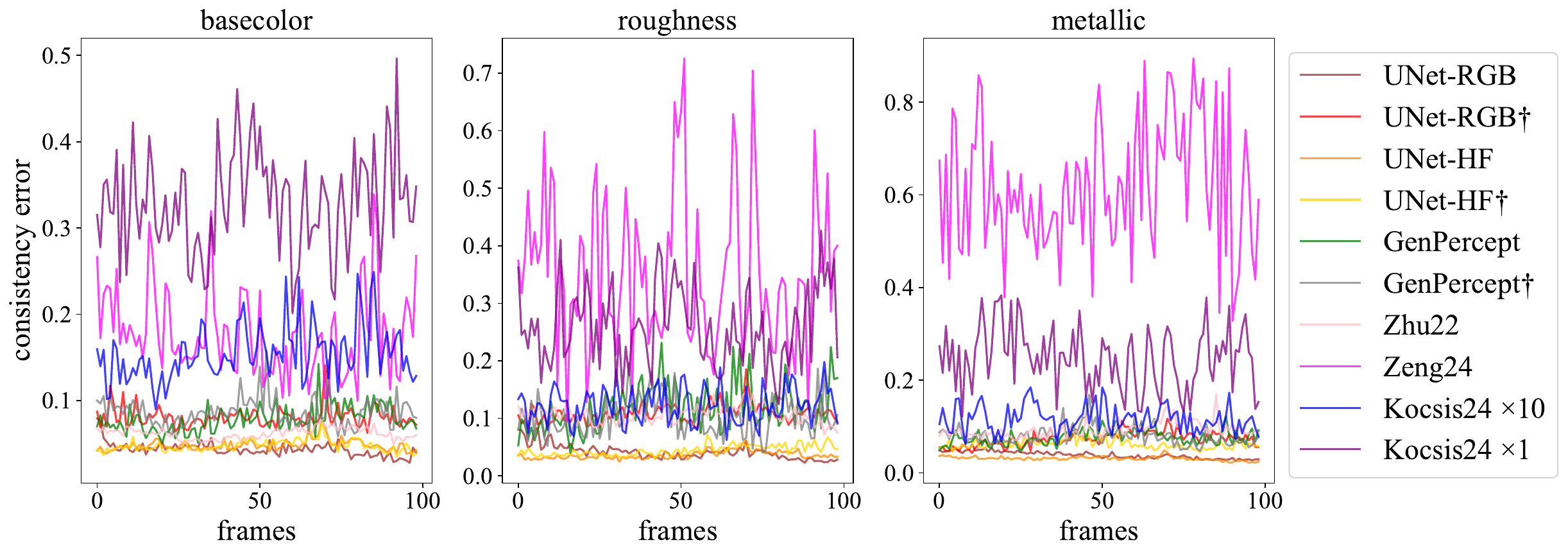}
      \caption{Stop-The-Pop metrics for the bedroom scene}
      \label{fig:stopthepop_bedroom}
\end{figure*}

\begin{figure*}[ht]
      \centering
      \includegraphics[width=\textwidth]{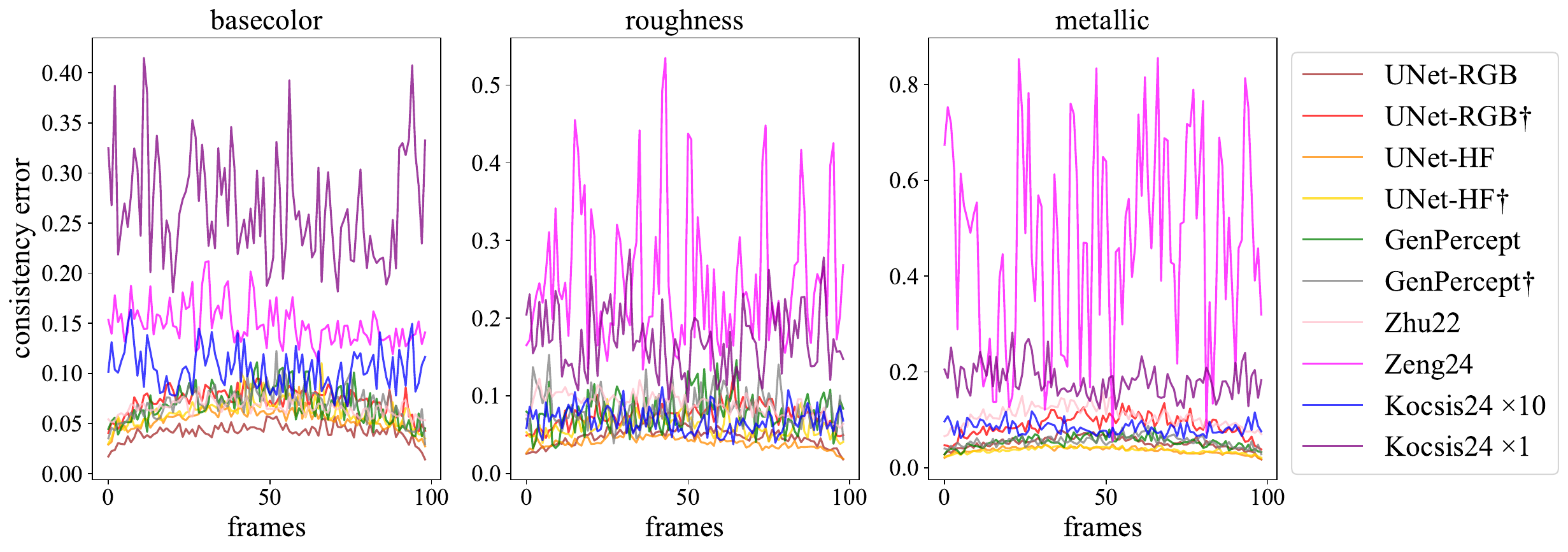}
      \caption{Stop-The-Pop metrics for the 'kitchen-003' scene}
      \label{fig:stopthepop_kitchen_003}
\end{figure*}

\begin{figure*}[ht]
      \centering
      \includegraphics[width=\textwidth]{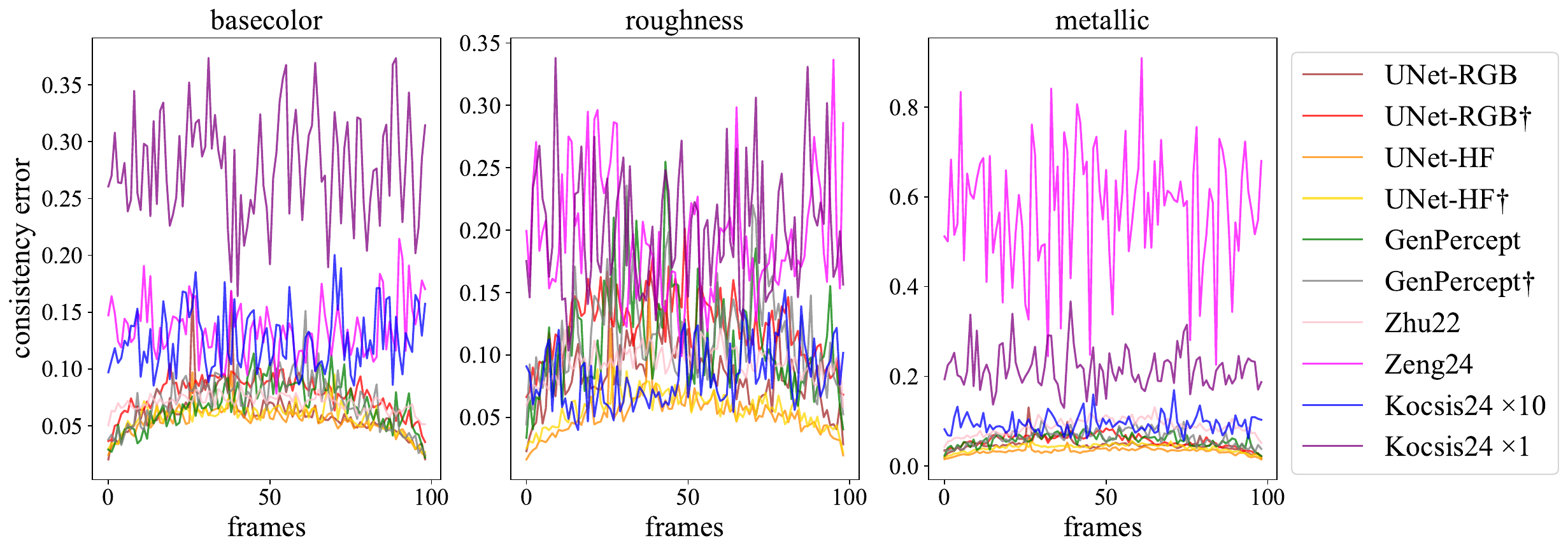}
      \caption{Stop-The-Pop metrics for the 'kitchen-005' scene}
      \label{fig:stopthepop_kitchen_005}
\end{figure*}

\begin{figure*}[ht]
      \centering
      \includegraphics[width=\textwidth]{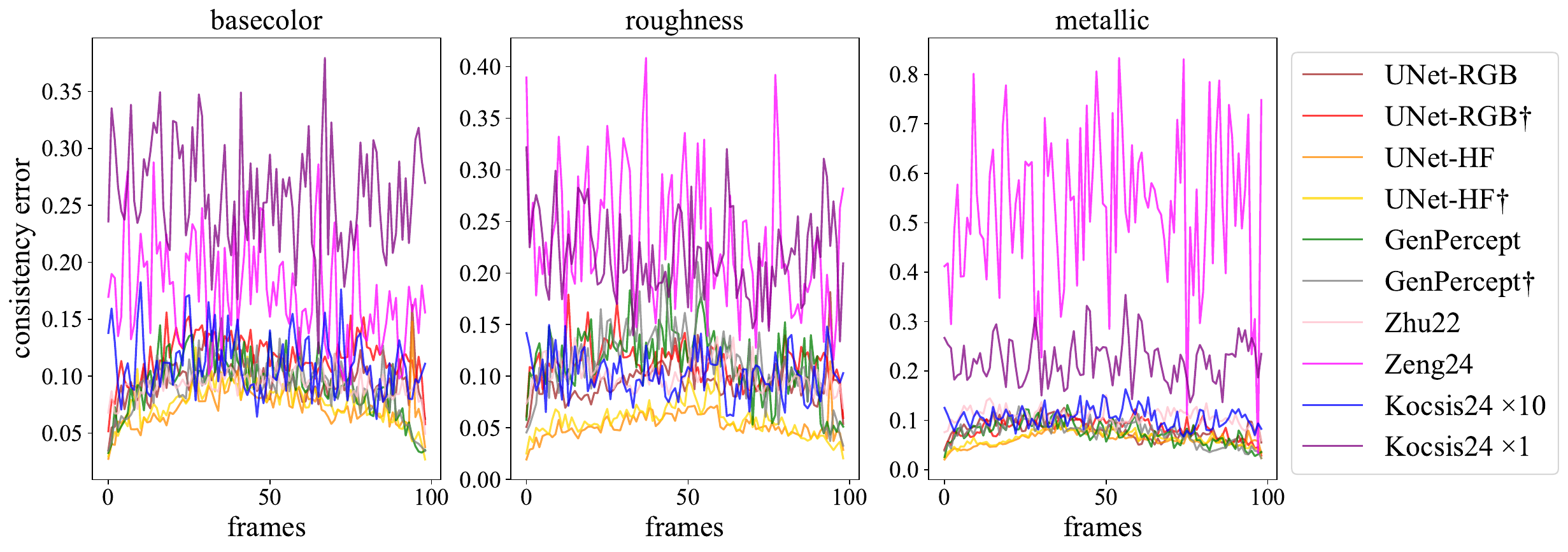}
      \caption{Stop-The-Pop metrics for the 'kitchen-007' scene}
      \label{fig:stopthepop_kitchen_007}
\end{figure*}

\begin{figure*}[ht]
      \centering
      \includegraphics[width=\textwidth]{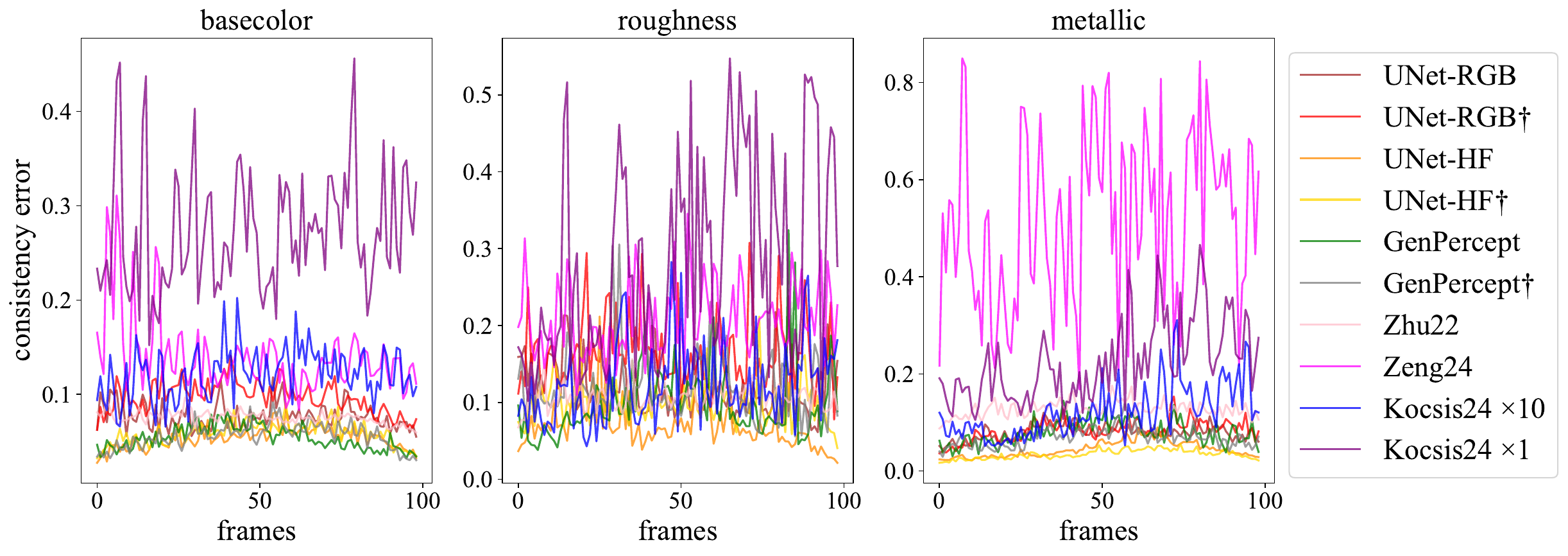}
      \caption{Stop-The-Pop metrics for the 'kitchen-010' scene}
      \label{fig:stopthepop_kitchen_010}
\end{figure*}

We provide the full Stop-The-Pop metric graphs Fig.\ref{fig:stopthepop_bedroom} to \ref{fig:stopthepop_kitchen_010} for the 5 scenes illustrated Fig.~\ref{fig:five_scenes}, which we used in the main article.

\section{Additional SVBRDF Predictions}

\begin{figure*}[ht]
      \centering
      \includegraphics[width=0.75\textwidth]{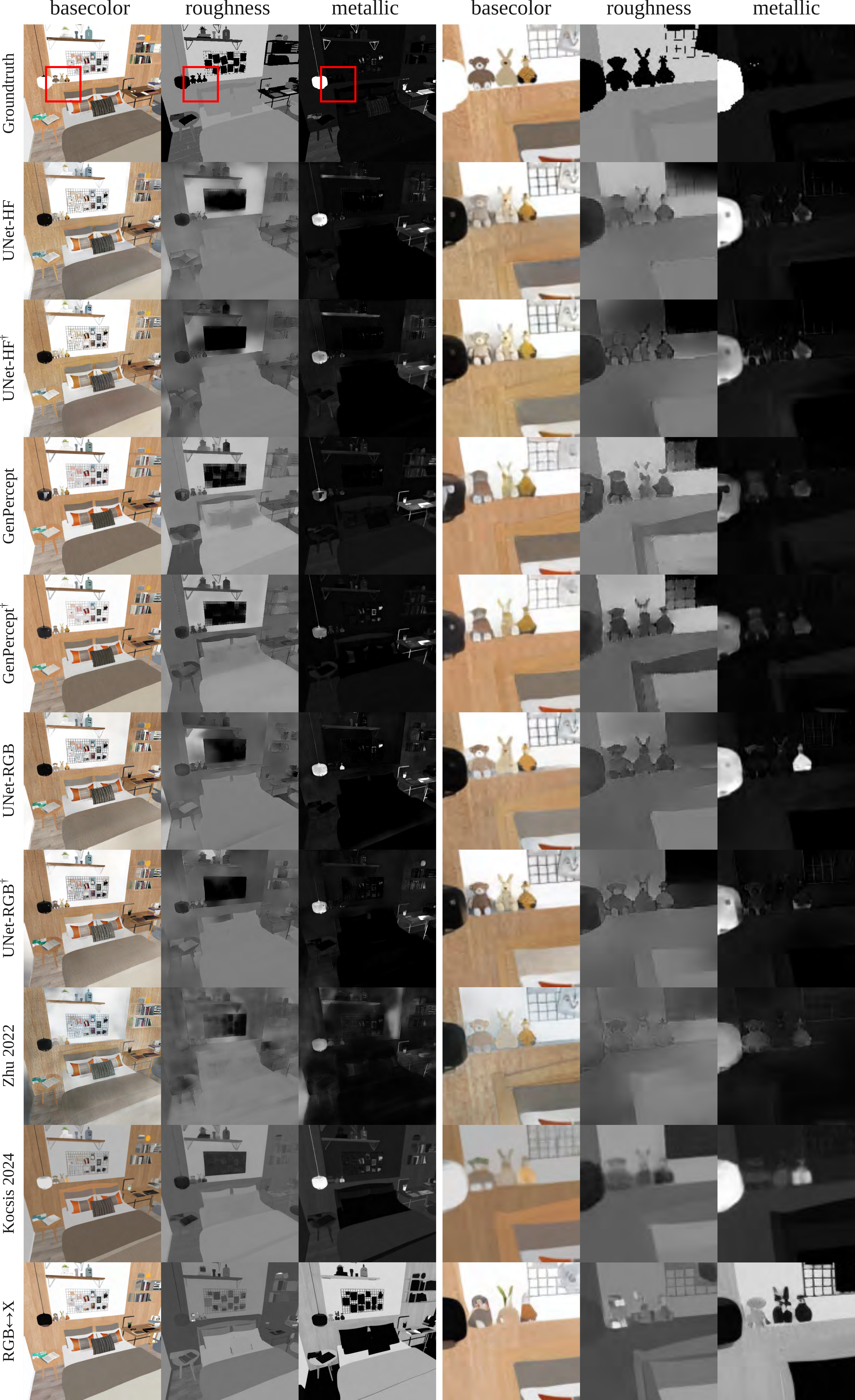}
      \caption{scene: L3D124S21ENDIMODSYQUI5NFSLUF3P3XG888; image 009}
      \label{fig:res1}
\end{figure*}

\begin{figure*}[ht]
      \centering
      \includegraphics[width=\textwidth]{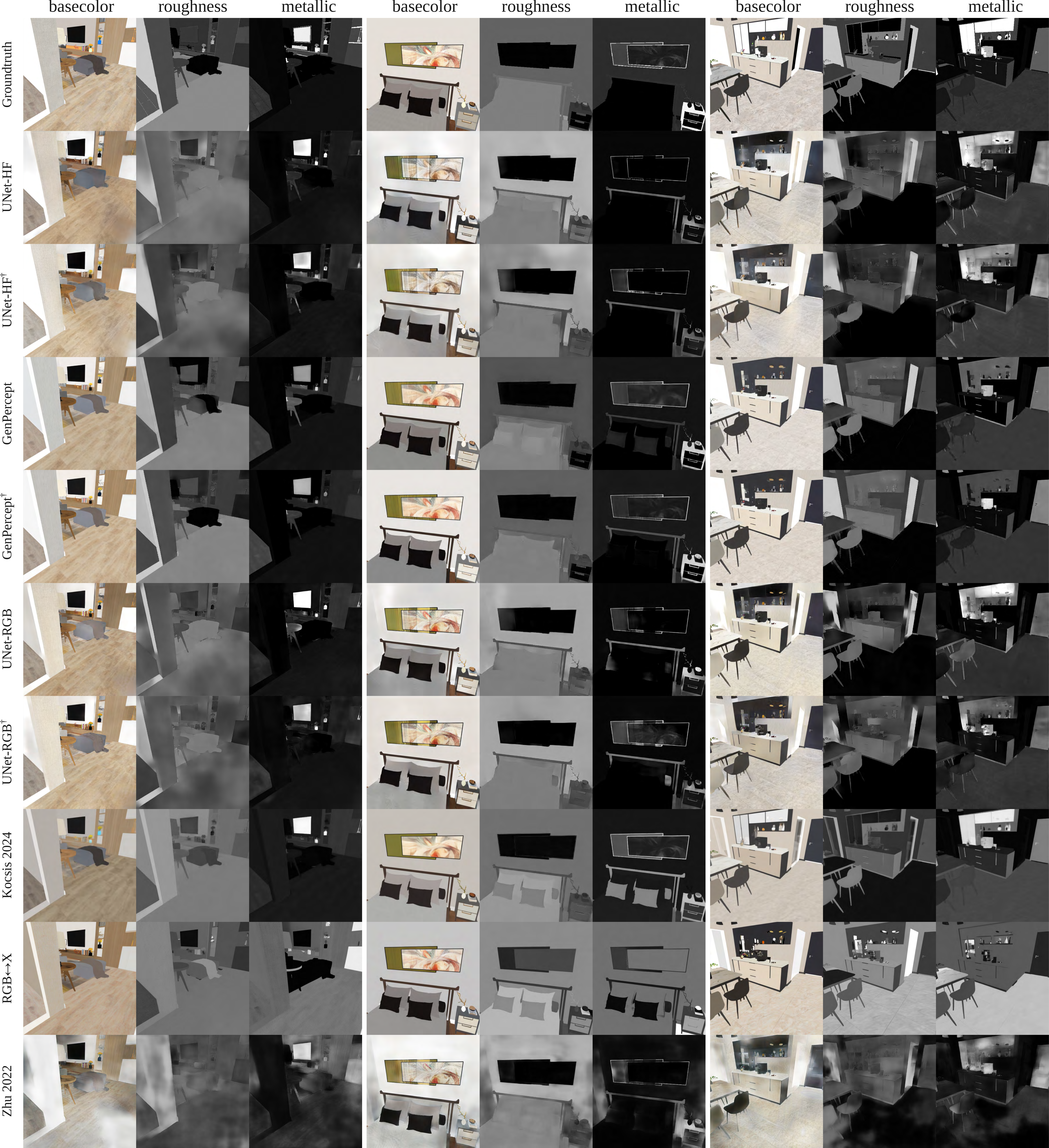}
      \caption{Scene names (from left to right): \\
            scene 1: L3D124S21ENDIDQO5UIUI5NFSLUF3P3XA888 - image 004 \\
            scene 2: L3D124S8ENDIDRGRTQUI5L7GLUF3P3WW888 - image 003 \\
            scene 3: L3D124S8ENDIMLAOPIUI5NYALUF3P3XS888 - image 004
      }
      \label{fig:resX}
\end{figure*}

In Figures~\ref{fig:res1} to \ref{fig:resX}, we illustrate more SVBRDF predictions results with all the methods we evaluate on 5 additional examples.

\begin{figure*}[ht]
      \centering
      \includegraphics[width=0.75\textwidth]{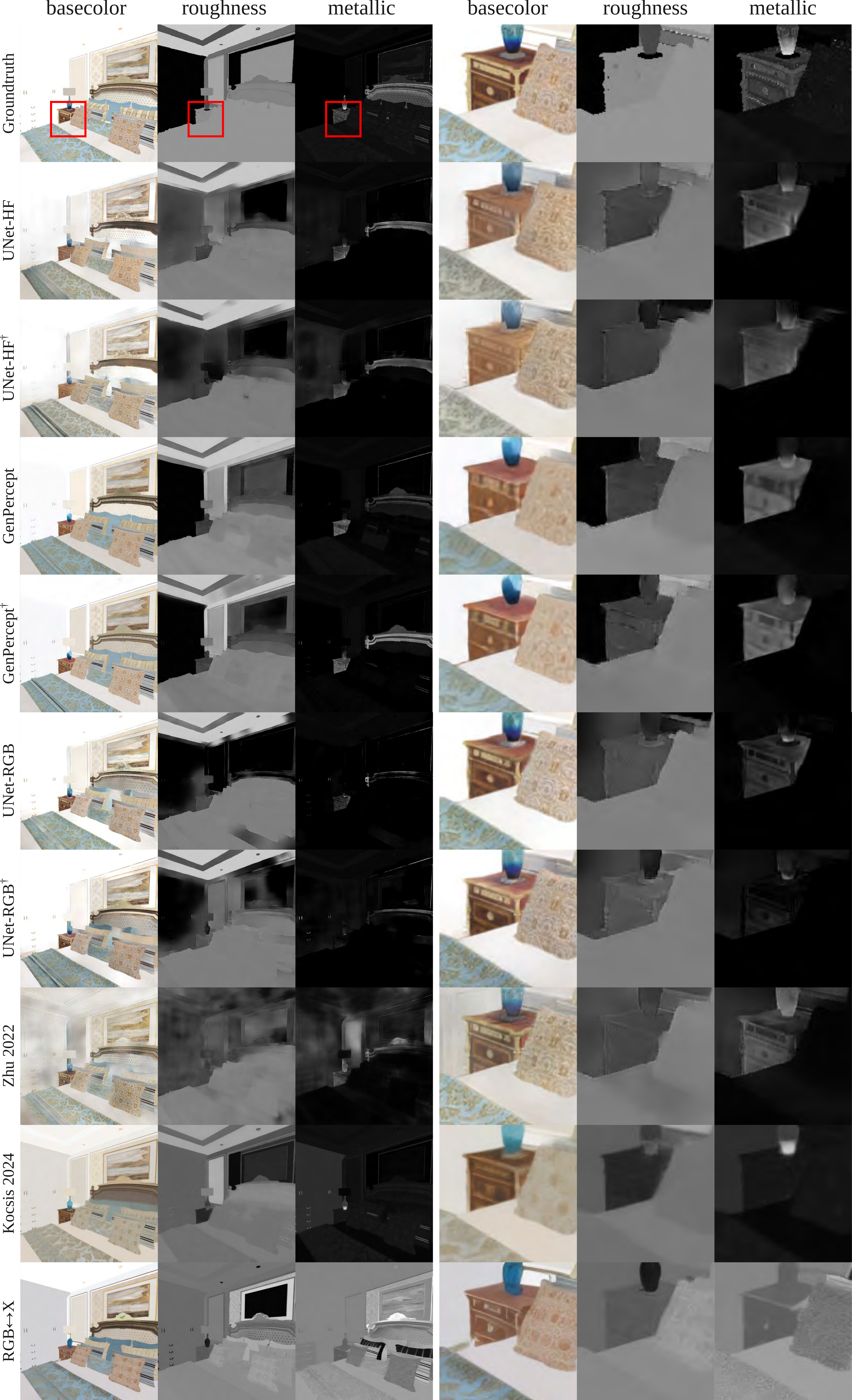}
      \caption{scene: L3D187S8ENDIDQLVHQUI5NYALUF3P3XK888; image 006}
\end{figure*}

\end{document}